\documentclass[sigconf, 10pt]{acmart}

\usepackage{multirow}
\usepackage{graphicx}
\usepackage{latexsym}
\usepackage{caption}
\usepackage{color}
\usepackage[noend]{algpseudocode}
\usepackage{url,epsfig,array}
\usepackage{leftidx}
\usepackage{amsmath}
\usepackage{epstopdf}
\usepackage{cases}
\usepackage{bm}
\usepackage{wasysym}

 \usepackage{bbding}

\usepackage{dsfont}
\usepackage{ifpdf}
\usepackage{epsfig}
\usepackage{amsfonts}

\usepackage{amssymb}
\usepackage{paralist}
\usepackage{comment}
\usepackage{xspace}
\usepackage{setspace}
\usepackage{balance}
\usepackage{subfigure}
\usepackage{threeparttable}

\usepackage{enumitem}
\usepackage{makecell}
\usepackage{newtxmath}
\usepackage{booktabs}
\usepackage{tikz}
\usepackage{listings}
\usepackage{courier} 

\usepackage{colortbl}

\usepackage[ruled,linesnumbered]{algorithm2e}

\usepackage{calc}

\usepackage{pgfplots}
\usepackage{colortbl}
\usepackage{xcolor}
\usepackage{adjustbox}
\pgfplotsset{compat=1.18}

\definecolor{dkgreen}{rgb}{0,0.6,0}
\definecolor{gray}{rgb}{0.5,0.5,0.5}
\definecolor{mauve}{rgb}{0.58,0,0.82}
\definecolor{dkyellow}{RGB}{220,130,90}
\definecolor{dkblue}{RGB}{0,60,170}
\definecolor{citeblue}{RGB}{0,0,128}
\definecolor{dkpurple}{RGB}{148,0,211}
\definecolor{myyellow}{RGB}{211,162,55}

\renewcommand\footnotetextcopyrightpermission[1]{}
\AtBeginDocument{%
  }

\setcopyright{acmlicensed}
\copyrightyear{2018}
\acmYear{2018}
\acmDOI{XXXXXXX.XXXXXXX}

\acmConference[ACM Conference '25]{Make sure to enter the correct
  conference title from your rights confirmation emai}{}{}
\acmISBN{978-1-4503-XXXX-X/18/06}

\acmSubmissionID{}

\def\ie{\textit{i.e.}\xspace}

\def\etal{\textit{et al.}\xspace}

\def\eg{\textit{e.g.}\xspace}

\begin{document}

\title{Efficient Federated Fine-Tuning of Large Language Models with Layer Dropout}

\author{Shilong Wang,~Jianchun Liu,~Hongli Xu,~Jiaming Yan,~Xianjun Gao}

\affiliation{%
    \institution{University of Science and Technology of China, China}
  \country{}
}

\renewcommand{\shortauthors}{Wang \etal}

\begin{abstract}
Fine-tuning plays a crucial role in enabling pre-trained LLMs to evolve from general language comprehension to task-specific expertise. To preserve user data privacy, federated fine-tuning is often employed and has emerged as the de facto paradigm. However, federated fine-tuning is prohibitively inefficient due to the tension between LLM complexity and the resource constraint of end devices, incurring unaffordable fine-tuning overhead. Existing literature primarily utilizes parameter-efficient fine-tuning techniques to mitigate communication costs, yet computational and memory burdens continue to pose significant challenges for developers. This work proposes DropPEFT, an innovative federated PEFT framework that employs a novel stochastic transformer layer dropout method, enabling devices to deactivate a considerable fraction of LLMs layers during training, thereby eliminating the associated computational load and memory footprint. In DropPEFT, a key challenge is the proper configuration of dropout ratios for layers, as overhead and training performance are highly sensitive to this setting. To address this challenge, we adaptively assign optimal dropout-ratio configurations to devices through an exploration-exploitation strategy, achieving efficient and effective fine-tuning. Extensive experiments show that DropPEFT can achieve a $1.3$--$6.3\times$ speedup in model convergence and a $40\%$--$67\%$ reduction in memory footprint compared to state-of-the-art methods.

\end{abstract}




\maketitle

\section{Introduction}\label{sec_introduction}
Modern NLP has undergone significant advancements across various domains in recent years \cite{vaswani2017attention, devlin2018bert, liu2019roberta, radford2019language, brown2020language, zhao2023survey, kasneci2023chatgpt}. Notable examples include healthcare diagnostics \cite{thirunavukarasu2023large}, sentiment analysis \cite{zhang2023enhancing}, and machine translation \cite{zhang2023prompting}. 
These breakthroughs are driven by large language models (LLMs), which are pre-trained on extensive public text corpora \cite{devlin2018bert, liu2019roberta, brown2020language, chowdhery2023palm, le2023bloom}. To fully realize the potential of NLP, these models are then
\emph{fine-tuned} on domain-specific datasets to optimize their performance for downstream tasks \cite{devlin2018bert, liu2019roberta, huang2022large}, such as question answering for mathematical problems.

Typically, task-specific data for fine-tuning, such as user messages or emails, are continuously generated by users across a variety of end devices (\eg, mobile and embedded devices). However, such data are privacy-sensitive in many cases, raising significant privacy concerns when collected for fine-tuning LLMs.
To address this issue, federated fine-tuning has emerged as the de facto methodology for privacy-aware LLM adaptation \cite{cai2023efficient, zhang2023fedpetuning, lin2021fednlp, xu2024fwdllm}. 

\noindent \textbf{Challenges.} 
Given the vast number of parameters in LLMs and the limited resources (\eg, communication bandwidth, computing power, and memory) available on end devices, the practical deployment of federated fine-tuning systems poses challenges, which are threefold. (1) Transmission of model updates between devices and the server incurs excessive \textbf{communication} time; (2) Expensive on-device \textbf{computation} for updating LLMs introduces significant delays; (3) Fine-tuning LLMs requires unaffordable \textbf{storage} space on end devices.
For instance, in each federated round, fine-tuning a relatively small LLM, \eg, GPT-2 \cite{radford2019language}, on each device even requires $12$ GB of network traffic and over $15$ petaFLOPs of computation. Considering that typical end devices, \eg, NVIDIA Jetson TX2, are capable of $<2$ TFLOPS of computational capability and $<100$ Mbps of bandwidth, one round of federated fine-tuning can take several hours, and the end-to-end convergence time can extend to hundreds of hours \cite{cai2023efficient}. Moreover, fine-tuning GPT-2 necessitates about $30$ GB of memory, which is impractical for most end devices, typically equipped with $<16$ GB of GPU memory \cite{kang2021benchmarking, magalhaes2023benchmarking, baller2021deepedgebench}.

\noindent \textbf{Status quo and limitations.} 
Recent research on federated fine-tuning primarily addresses the communication issue, with federated parameter-efficient fine-tuning (PEFT) \cite{cai2023efficient, zhang2023fedpetuning, jiang2023low, cho2023heterogeneous} emerging as the dominant strategy. Specifically, PEFT inserts lightweight trainable modules into the LLM while keeping the base model frozen. During communication rounds, only parameter updates from the added modules are shared between devices and the server. Since the size of these modules is typically less than 5\% of the base LLM, network traffic for federated fine-tuning is significantly reduced.
However, the fine-tuning overhead remains significant for developers with the PEFT methods. This is because PEFT cannot fundamentally address the computation and storage challenges, which act as primary bottlenecks of federated fine-tuning. As demonstrated in \S\ref{subsec:peft}, computation time of PEFT remains substantially high, \eg, about one hour per round when fine-tuning a 1.5B LLM even using a powerful end device, which is $99\times$ longer than communication time.
Moreover, PEFT provides limited memory savings and requires about 20 GB of GPU memory for fine-tuning the 1.5B model. Consequently, it is impractical to deploy popular LLMs on real-world end devices.

\noindent \textbf{Root causes of limitations.} The computation and storage bottlenecks in PEFT stem from its \emph{additive} design philosophy: it grafts new parameters onto each transformer layer in the LLM rather than compressing these layers. First, the architectural choice of PEFT imposes an unaffordable computational burden. During fine-tuning, inputs must propagate through \emph{all} transformer layers, even those with frozen parameters. While PEFT accelerates the backward pass during fine-tuning, the computational graph of the forward pass for the frozen base model remains intact. The majority of FLOPs in the forward pass originate from the frozen base model, which PEFT does not skip. Consequently, PEFT fails to optimize the forward pass, whose computational load remains nearly identical to that of full fine-tuning (FFT) without any parameters frozen, 
accounting for about 45\% of the total on-device computation time (\S \ref{subsec:limitations}). Second, fine-tuning LLMs necessitates extensive GPU memory to store intermediate results, \ie, activations \cite{korthikanti2023reducing}. This is because gradients used to update PEFT modules depend recursively on activations from preceding transformer layers. For example, computing gradients for a PEFT module at layer $L$ requires cached activations from layer $(L-1)$. Thus, \emph{all} layers must retain intermediate activations to support gradient calculations.
Consequently, the activations generated during the forward pass take up most of the memory usage ($>79\%$), which cannot be eliminated by PEFT (\S \ref{subsec:limitations}).

\noindent \textbf{Our solution.} We thereby present an efficient framework called DropPEFT, which introduces a novel layer-wise optimization strategy, \ie, stochastic transformer layer dropout (STLD), as the key building block to enhance PEFT. Unlike conventional PEFT methods that retain all transformer layers during training, DropPEFT dynamically identifies and skips certain computationally costly layers in both forward and backward passes.
The intuition behind DropPEFT is inspired by the stochastic depth method \cite{huang2016deep}, which is used to train ResNets \cite{he2016deep} efficiently by shortening the network depth.
Specifically, during each training batch, participating devices in DropPEFT fine-tune the LLM by stochastically deactivating subsets of transformer layers according to certain probabilities (\ie, dropout rates). Inputs propagate only through activated layers, while deactivated layers drop out both of the forward and backward passes.
Critically, this dropout of deactivated layers is temporary and dynamic: a layer deactivated in one batch may be reactivated in subsequent batches, ensuring all layers in the LLM contribute cumulatively over time. By STLD, DropPEFT eliminates both computations and activations for deactivated layers. Consequently, the memory usage and training time required for fine-tuning are significantly reduced.

However, unleashing DropPEFT's full potential presents a key technical challenge: \emph{how to design an effective strategy to assign an appropriate dropout rate to each transformer layer.} This drop-rate configuration govern the trade-off between fine-tuning efficiency and model fidelity. Overly aggressive dropout introduces a risk of degrading the model performance, while overly conservative dropout rates squander computational and memory resources. Compounding this, the importance of a layer varies with its position in the LLM, leading to different suitable dropout rates for layers at various positions.
The choice is also dynamic: the favorable drop-rate configuration drifts over time and devices, depending on the learning progress and changing device resources. 
To address this, DropPEFT employs an online exploration-exploitation strategy that dynamically optimizes dropout rates based on real-time rewards, quantified as model accuracy gains per unit wall-clock time. This adaptive mechanism evaluates candidate configurations through a multi-armed bandit framework, prioritizing high-reward dropout policies while maintaining exploratory diversity. By continuously adapting to demands of devices and learning phase, DropPEFT achieves efficient and effective fine-tuning.

\noindent \textbf{Contributions.} 
Overall, we make the following contributions in this paper:
\begin{itemize}[leftmargin=*]
    \item We analyze the limitations of PEFT in computation and storage efficiency. To make federated fine-tuning practical for end devices, we propose DropPEFT to enhance PEFT with a novel stochastic transformer layer dropout method.
    \item We identify the challenges of the drop-rate configuration in DropPEFT, then design an online exploration-exploitation algorithm to determine the optimal configuration for efficient and effective layer dropout.
    \item Considering the practical issue of statistical heterogeneity in federated fine-tuning, we extend DropPEFT by incorporating a personalized layer sharing method (\S\ref{sec:personalized}), which improves DropPEFT's adaptability to heterogeneous data.
    \item Through experiments on real hardware, we demonstrate that DropPEFT outperforms state-of-the-art solutions significantly across various datasets and models.

\end{itemize}

\section{Background and Motivation}\label{sec:background}

\subsection{Challenges in Federated Fine-Tuning}\label{subsec:challenges}
Federated fine-tuning of LLMs, \eg, DeBERTaV2-xxlarge (1.5B) \cite{he2020deberta}, on typical end devices such as Jetson TX2 \cite{tx2} and NX \cite{nx} faces formidable challenges due to the stark mismatch between LLM’s significant resource demands and devices’ limited capabilities. For instance, fine-tuning DeBERTaV2 on a Jetson TX2’s 256-core GPU takes orders of magnitude longer than on server-grade GPUs, while energy constraints on battery-powered devices render sustained training impractical. Moreover, communication overhead compounds these issues. Transmitting 1.5B parameter updates over a typical 40 Mbps uplink/downlink requires over 40 minutes per communication round, making frequent synchronization infeasible. Storage limitations further hinder deployment, as training a 1.5B-parameter LLM necessitates storing over 25 GB of model weights and intermediate results (\S \ref{subsec:limitations}), which exceeds the memory capacity of most devices. These constraints underscore a fundamental discrepancy: federated fine-tuning demands high computational, communication, and memory resources, whereas most end devices lack the hardware capabilities to meet these requirements.

\subsection{PEFT: Benefits and Limitations}\label{subsec:peft}

\noindent \textbf{PEFT and its benefits.} Recent attempts incorporate PEFT techniques (\eg, LoRA \cite{zhang2023fedpetuning, cho2024heterogeneous} and Adapter \cite{zhang2023fedpetuning, cai2023efficient}) into federated fine-tuning and achieves significant improvements in mitigating communication costs (\ie, delays and network traffic).
Specifically, they insert small, additional modules (\eg, adapters or low-rank matrices) into the transformer layers of the LLM and keep the the original model frozen. During fine-tuning, only updates from these modules, not the entire LLM, are transmitted between devices and the server.
Since PEFT introduces only a small number of additional parameters relative to the frozen ones, it alleviates the majority ($>95\%$) of communication overhead.

\noindent \textbf{Limitations.} PEFT is at very early stage towards practical federated fine-tuning, as it cannot fully address many other issues such as excessive computational burden and memory footprint. To reveal these limitations, we fine-tune DeBERTaV2-xxlarge on the MNLI dataset \cite{wang2018glue} with Jetson AGX \cite{agx}, a high-performance embedded device developed by NVIDIA. As illustrated in Table \ref{tab:memory_time}, while communication time has decreased by over $99\%$ with the PEFT methods, computation time remains substantially high (\eg, about one hour per round on each device), accounting for more than 99\% of the total fine-tuning time. When fine-tuning the LLM on weaker devices, \eg, TX2 and NX, the computation time could be further increase. Moreover, PEFT provides limited memory savings (approximately $30\%$), making it impractical to deploy popular LLMs on real-world end devices with $< 16$ GB of memory. More severely, even only a portion of that memory on end devices can be allocated for fine-tuning tasks without compromising user experience \cite{lebeck2020end, li2017optimizing}.

\begin{table}[t]
    \includegraphics[width=8.5cm]{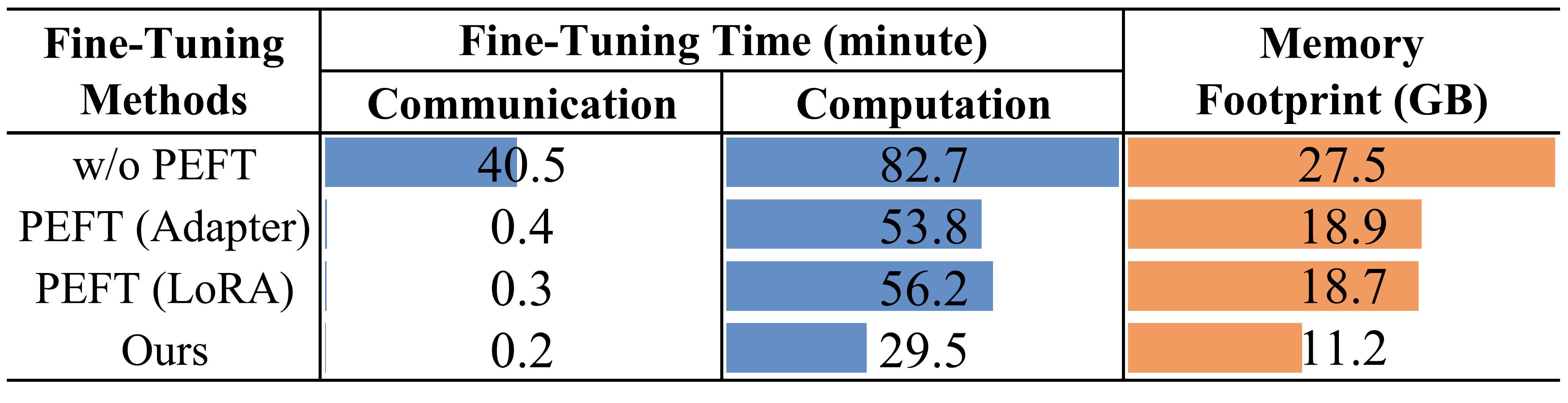}
\caption{Communication (both downlink/uplink), computation, and storage overhead during a single round on each device. Bandwidth: 40 Mbps.}
\label{tab:memory_time}
\vspace{-0.5cm}
\end{table}

\subsection{Root Causes of Limitations}\label{subsec:limitations}
In this subsection, we delve into a comprehensive analysis on why PEFT cannot effectively address the issues of computation and memory usage.

\begin{figure}[t]\centering
    \includegraphics[width=1.0\linewidth]{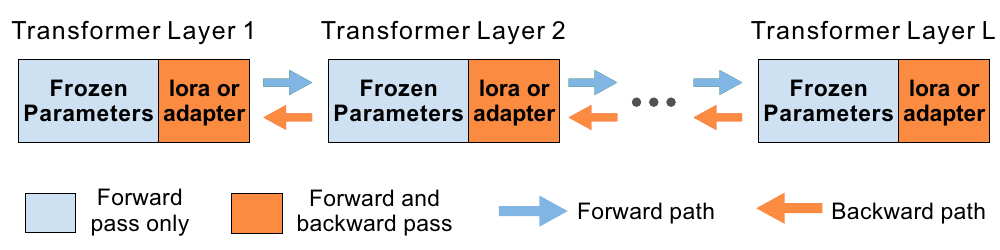}
    \caption{Illustration of forward and backward passes in parameter-efficient fine-tuning.}\label{fig:computing_path}
\end{figure}

\begin{figure}[t]\centering
    \includegraphics[width=0.985\linewidth]{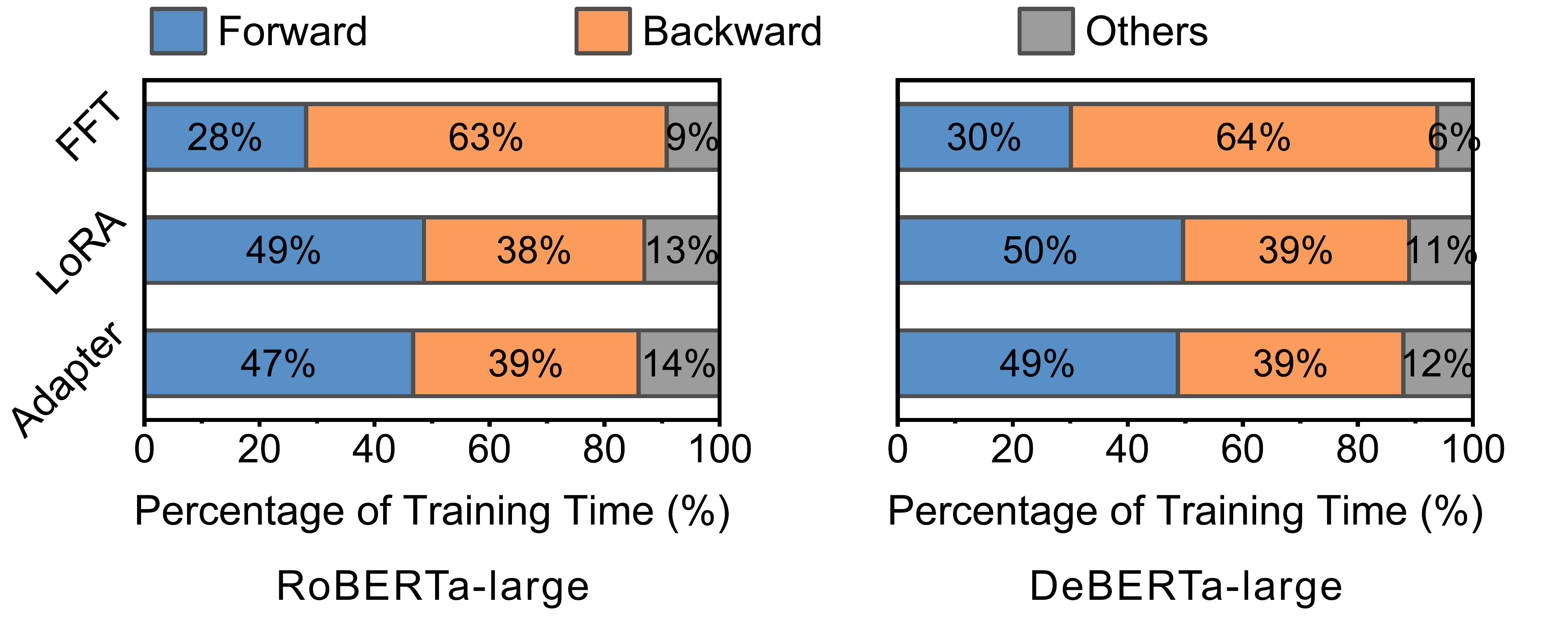}
    \vspace{-2mm}
    \caption{Breakdown of computation time.}\label{fig:breakdown-time}
\end{figure}

\noindent \textbf{PEFT leaves the forward pass computationally costly.} 
The computational overhead in LLM fine-tuning is dominated by the forward and backward passes of the LLM’s backbone \cite{dettmers2023qlora}. We observe that traditional PEFT methods only improve the efficiency of the backward pass but fail to reduce the computational demands of the forward pass. As shown in Figure \ref{fig:computing_path}, while PEFT freezes the parameters in the base LLM to avoid gradient calculations during the backward pass, it does not alter the computational graph in the forward pass, leaving the original forward path in the base LLM intact. Common LLMs, such as LLaMA-7B, contain billions of parameters accessed during the forward pass, making it computationally demanding in terms of training time. Moreover, PEFT introduces additional modules, which further increase the computation complexity for the forward pass. To illustrate this, we fine-tune RoBERTa-large \cite{liu2019roberta} and DeBERTa-large \cite{he2021debertav3} on the MNLI dataset using an A6000 GPU. During fine-tuning, we record time for the forward pass, backward pass, and other steps (\eg, data loading and optimizer stepping), respectively. As shown in Figure \ref{fig:breakdown-time}, although PEFT methods reduce the backward pass time, they fail to address the forward pass overhead. Consequently, the forward pass accounts for near 50\% of the total computation time, emerging as a primary bottleneck in LLM fine-tuning.

\begin{figure}[t]\centering
    \includegraphics[width=1.0\linewidth]{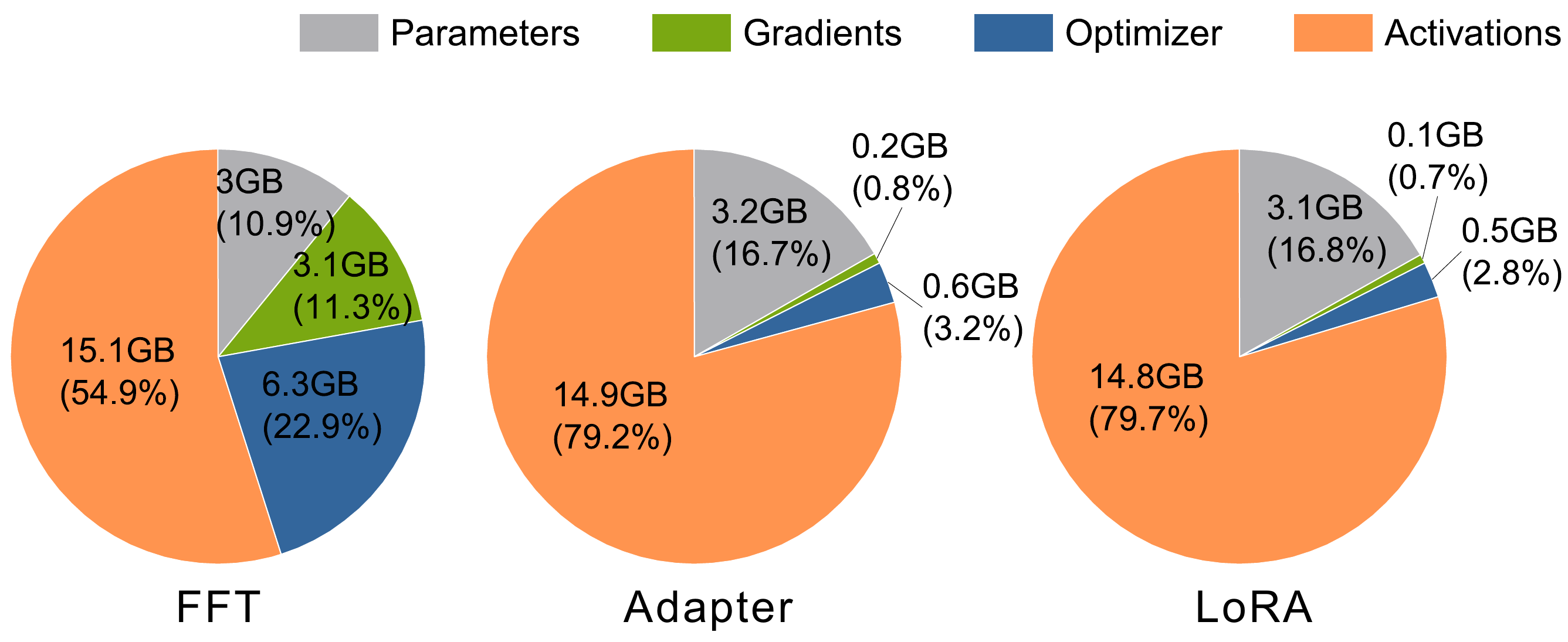}
    \caption{Breakdown of GPU memory footprint with a batch size of 16, maximum sequence length of 256 \cite{MicrosoftDeberta}, and the commonly used Adamw optimizer (BF16 numerical
format) \cite{loshchilov2017decoupled}.}\label{fig:breakdown_memory}
\end{figure}

\noindent \textbf{Memory overhead of PEFT is significant due to the large size of activations.} To better understand the gap between the memory needed in existing PEFT techniques and the memory available on devices, we profile the memory requirements to fine-tune DeBERTaV2-xxlarge on MNLI. As shown in Figure \ref{fig:breakdown_memory}, the memory footprint for FFT includes storing the LLM’s parameters (10.9\%, grey), activations (54.9\%, orange), gradients (11.3\%, green), and optimizer states (22.9\%, blue). Even after the PEFT methods frozen the LLM parameters to alleviate the memory usage for gradients and optimizer states, the overall demands is still significant due to the \emph{unreduced} activations size (80\% of the total). The reason is that, activations are intermediate outputs of each layer during the forward pass, required for computing gradients in the backward pass. 
Their memory footprint is mainly determined by the LLM depth $L$, \ie, the number of transformer layers \cite{korthikanti2023reducing}.
Since PEFT 
fully preserves the original LLM's architecture, activations from all layers must still be stored. Consequently, there remains a $1.58\sim2.37\times$ gap between the memory required for PEFT and the memory available on commonly used end devices, \eg, 8GB for TX2 and 16GB for NX.


\subsection{Opportunities}
Despite the limitations of existing PEFT techniques, we also identify opportunities to develop efficient federated fine-tuning frameworks. Our analysis in \S\ref{subsec:limitations} demonstrates that the massive scale of LLMs, exemplified by DeBERTa-xxlarge with its 48 transformer layers, inherently amplifies fine-tuning costs. Current PEFT approaches exacerbate this issue by typically updating trainable parameters across most or all transformer layers, rendering them impractical for resource-constrained environments. To overcome this barrier, one intuitive opportunity might involve cutting down the size of LLMs by pruning the model depth. For example, halving the layers in DeBERTaV2 to $24$ could theoretically slash training latency and memory usage by about $50\%$.
However, such a brute-force transformer-layer reduction risks disrupting the capabilities of LLMs. Neural scaling laws \cite{sorscher2022beyond, kaplan2020scaling} confirm that LLM performance scales predictably with depth, as deeper architectures capture more comprehensive semantic information. Therefore, immediate gains in efficiency by removing transformer layers come at the cost of irreversible damage to LLM capability. This prompts a pivotal question: \emph{Can PEFT methods be redesigned to strategically alleviate computational and memory burdens of certain layers without compromising the intrinsic fidelity of LLMs?}

\section{Design of DropPEFT}\label{sec:design}

\subsection{Overview}
In this work, we propose DropPEFT, an enhanced federated PEFT framework that improves efficiency through a novel \emph{stochastic transformer layer dropout (STLD)} technique. DropPEFT employs
a similar training process as traditional federated PEFT framework but mainly differs on the local fine-tuning paradigm. In each global round, the central server first sends the latest PEFT modules to available devices. Then each device inserts these modules into corresponding transformer layers of the local LLM, which is fine-tuned on the device's local training data by STLD for several mini-batches. The key idea behind STLD roots in the seemingly paradoxical insight that maximal LLM depth is essential for superior model performance, but a short network is beneficial for efficient fine-tuning. At its core, during local fine-tuning, STLD dynamically shortens the active network depth by randomly skipping a substantial fraction of transformer layers independently for each mini-batch, while retaining the LLM's full original architecture post-training. Finally, the devices upload the updates of PEFT modules from all transformer layers to the server for global aggregation.

\noindent \textbf{Advantages of DropPEFT by design.} 
First, by deactivating selected layers, STLD circumvents both forward and backward computations for those layers, thereby significantly reducing computational overhead. Second, the omission of these layers during computation eliminates the need to store intermediate activations, gradients, and optimizer states, thus alleviating memory constraints. Finally, unlike permanent pruning techniques, STLD retains all layers after training, ensuring that the LLM’s representational capacity is maximally preserved.

\subsection{Stochastic Transformer Layer Dropout} \label{sec:layer_dropout}
We begin by reviewing the architecture of LLMs, which naturally lends itself to the design of STLD. 
LLMs are constructed by stacking $L$ structurally identical transformer layers, each acting as a fundamental building block. Between adjacent layers, hidden states serve as intermediate representations of the input sequences. More formally, each transformer layer $l$ processes the hidden states $H_l$ to yield higher-level representations $H_{l+1}$:
\begin{equation} \label{eq:transformer_layer}
    H_{l+1} = TransforBlock_{l}(H_l),
\end{equation}
where transformation function $TransforBlock_{l}(\cdot)$ 
encapsulates all computations within transformer layer $l$, including the self-attention and feed-forward processes \cite{vaswani2017attention}. In the context of PEFT methods, $TransforBlock_{l}(\cdot)$ further integrates computations associated with inserted PEFT modules (\eg, adapters or low-rank matrices). 

\begin{figure}[t]\centering
    \includegraphics[width=1.0\linewidth]{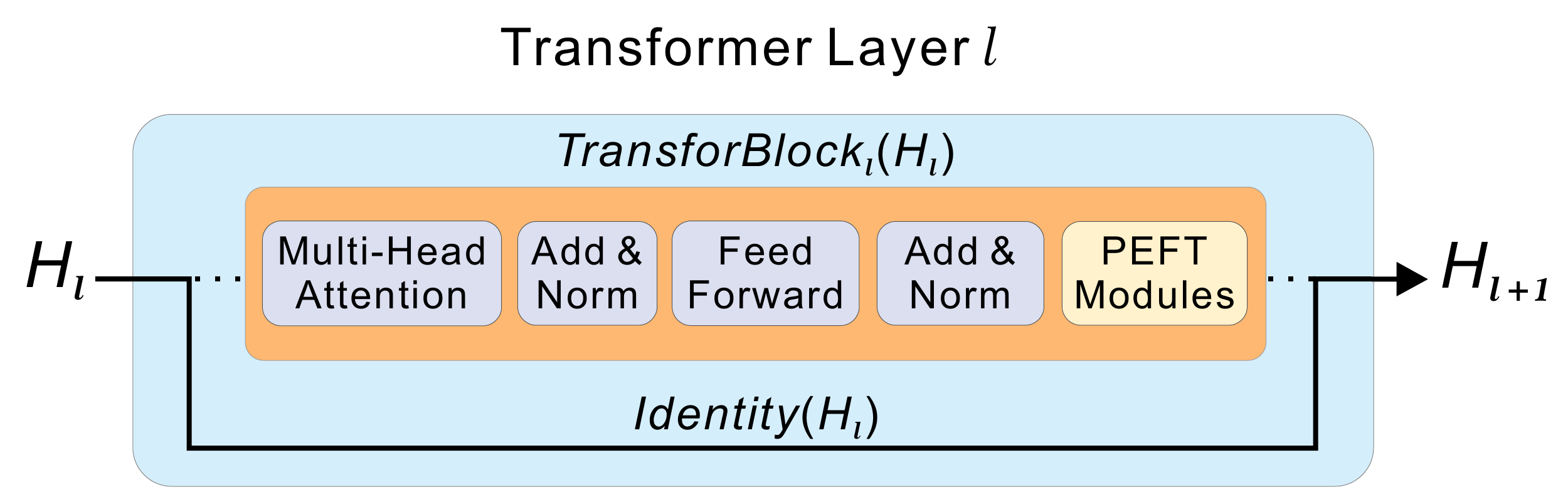}
    \caption{A close look at the deactivated layer $l$.}\label{fig:dropped-layer}
\end{figure}


The core of STLD is to construct and train shorter subnetworks by randomly deactivating certain transformer layers for each mini-batch. 
Figures \ref{fig:dropped-layer} offers a schematic view of a deactivated layer. If a transformer layer $l$ is chosen to be deactivated, the hidden states $H_{l}$ are passed through an identity function rather than $TransforBlock_{l}(\cdot)$:
\begin{equation}
    H_{l+1} = Identity(H_{l})
\end{equation}
where $Identity(\cdot)$ simply returns the input unchanged.
If the layer $l$ remains active, the original transformation in Equation~\eqref{eq:transformer_layer} is applied as usual. To model the process of selecting which layers to deactivate, we introduce a binary random variable $d_l \in \{0,1\}$ for each layer $l$. Here, $d_l = 1$ denotes that the layer is deactivated, while $d_l = 0$ indicates the layer is activated:
\begin{equation}\label{eq:layer-dropout}
    H_{l+1} = (1-d_l) \cdot TransforBlock_l(H_l) + d_l \cdot Identity(H_l)
\end{equation}
where the probability of $d_l = 1$, \ie, the dropout rate for layer $l$, is $P_l \in \left[0, 1\right)$. By assigning distinct dropout rates to each layer, we ensure that all layers have opportunities to contribute to fine-tuning across mini-batches over time.


\noindent \textbf{The rationales behind STLD.} Training with STLD can be viewed as training an ensemble of subnetworks that share the parameters in the overlapping layers. Specifically, for a LLM with $L$ transformer layers, there are $2^L$ possible subnetworks, each corresponding to a different combination of active and inactive layers. In each mini-batch, one of these subnetworks is selected for updating. This encourages the model to generalize better by preventing over-reliance on any particular subnetwork. Unlike model pruning, which permanently removes layers from the model, STLD can retain a full model at inference time by keeping all transformer layers active.
Therefore, it achieves efficiency during training while still benefiting from the representational richness of the complete architecture during inference. Figure~\ref{code:stochastic_layer_dropout} presents a code example of STLD. Notably, STLD accelerates training and reduces memory usages by reducing the number of active layers without requiring any specialized hardware or software kernels. 

\lstdefinelanguage{mypython}{        
  sensitive=true,
  keywords={def,class,return,if,else,import,as,for,in, self, range}, 
  morekeywords=[2]{int,str,bool,None}, 
  otherkeywords={->},          
  keywordstyle=\color{dkblue}\bfseries,  
  keywordstyle=[2]\color{magenta}\bfseries,  
  comment=[l]{\#},        
  morecomment=[s]{/*}{*/}, 
  commentstyle=\color{dkgreen}\scriptsize, 
  stringstyle=\color{mauve},           
}
\lstset{
    language=mypython,
    frame=b,                        
    xleftmargin=6pt, xrightmargin=3pt, 
    aboveskip=2mm,                     
    belowskip=2mm,                     
    basicstyle=\fontsize{7pt}{9pt}\ttfamily,  
    numbers=left,   
    numbersep=5pt,  
    numberstyle=\scriptsize\color{gray}, 
    showstringspaces=false,
    columns=flexible,
    breaklines=true,              
    breakatwhitespace=true,
    tabsize=4,                    
    extendedchars=false,          
    escapeinside={(*@}{@*)}    
}


\begin{figure}
\centering
\begin{lstlisting}
import random
import torch.nn as nn

class LLM(PreTrainedModel):              # e,g, LlamaModel
    def __init__(self, config):
        ...                    # omit codes in Transformers
        self.layers = nn.ModuleList(  # transformer layers
            [LLMLayer(config, layer_idx) for layer_idx in range(config.num_hidden_layers)]
        )
    
    def forward(self, (*@\footnotesize\textcolor{magenta}{drop\_rates}@*), hidden_states, ...):
        ...
        # for each transformer layer
        for i, layer in enumerate(self.layers):
            # stochastic transformer layer dropout
            (*@\footnotesize\textbf{\textcolor{magenta}{if self}}\textcolor{magenta}{.training:}@*)
                (*@\footnotesize\textcolor{magenta}{drop = random.random()}@*)
                (*@\footnotesize\textbf{\textcolor{magenta}{if}} \textcolor{magenta}{drop < drop\_rates[i]:}@*)
                    (*@\footnotesize\textbf{\textcolor{magenta}{continue}}@*)
            # transformation function of transformer layer
            hidden_states = layer(hidden_states, ...)
            ...
        ...
\end{lstlisting}
\captionof{figure}{Code snippet of STLD built atop the Transformers library \cite{Transformers}.
The codes we inserted into Transformers are marked in \textcolor{magenta}{red}.} \label{code:stochastic_layer_dropout}
\vspace{-3mm}
\end{figure}

\noindent \textbf{Computation and memory overhead analysis.} For each mini-batch, the number of layers \emph{activated}, denoted as $\widetilde{L}$, is a random variable. Under the assumption of independent deactivations across layers, the expected number of $\widetilde{L}$ is:
\begin{equation}
    \mathbb{E}(\widetilde{L}) = \sum_{l=1}^{L}(1-P_{l})
\end{equation}
where $P_l$ is the dropout rate for layer $l$.
Thus, rather than consistently training all \(L\) layers, we train a subnetwork with an average number of $\mathbb{E}(\widetilde{L})$ layers for each batch. This reduced model depth directly cuts down not only the chain of the forward pass but also gradient computations of the backward pass. Besides, intermediate activations, gradients, and optimizer states associated with those deactivated layers are fully eliminated.
Consequently, fine-tuning with STLD can reduce both the computation delay and memory usage by approximately $[L - \mathbb{E}(\widetilde{L})] / L$, compared to standard PEFT with all layers active. The practical experimental results are consistent with this analysis (see \S\ref{subsec:run_time}).

\subsection{Configurator for dropout rates}\label{sec:configurator}
\begin{figure}[t]\centering
    \begin{minipage}[t]{0.5\linewidth}\centering
        \subfigure[Impact of the dropout rate degree (each layer has the same dropout rate).]{\centering
                \label{fig:impact-average-drop-rate}
                \includegraphics[width=0.99\linewidth]{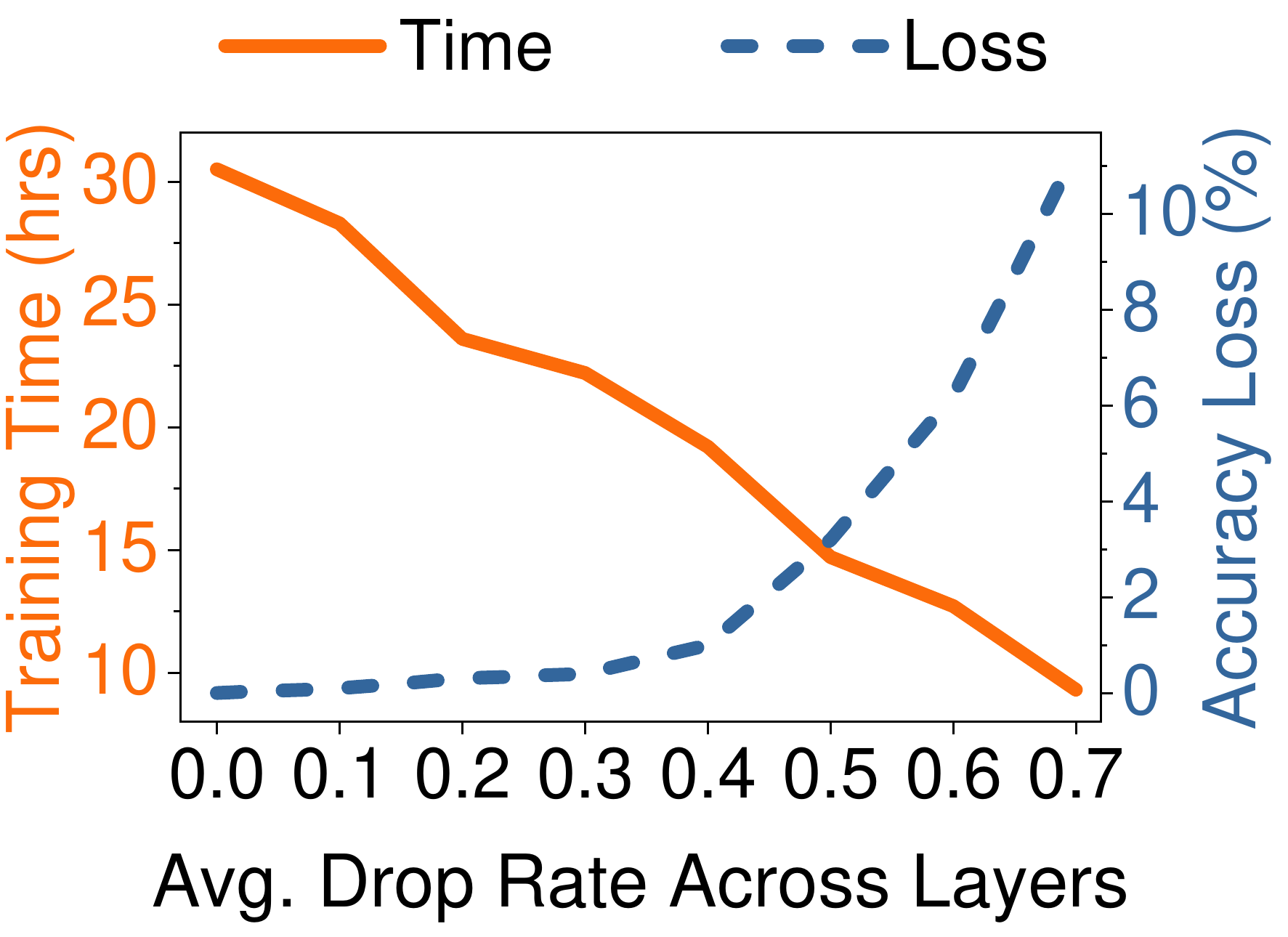}
            }
    \end{minipage}
    \begin{minipage}[t]{0.49\linewidth}\centering
        \subfigure[Impact of the dropout rate distribution across layers (the average dropout rate is 0.5).]{\centering
                \label{fig:impact-drop-rate-distribution}
                \includegraphics[width=0.835\linewidth]{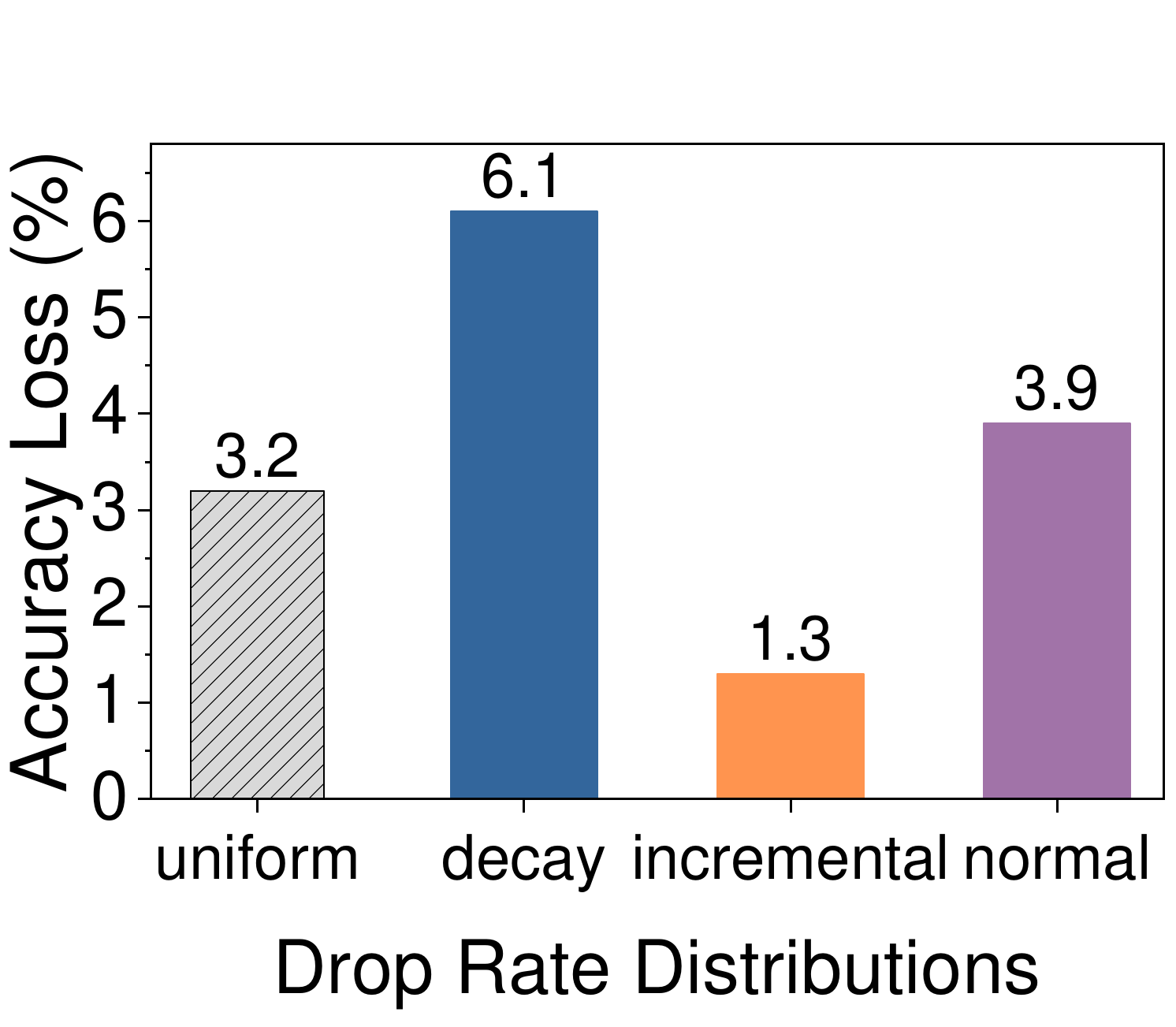}
            }
    \end{minipage}
    \vspace{-5mm}
    \caption{Training performance under various layer dropout configurations. Model \& Dataset: RoBERTa-large and MNLI.
    In Figure \ref{fig:impact-drop-rate-distribution}, the \emph{uniform} distribution sets $P_l=0.5$ uniformly for each layer $l$; the \emph{decay} distribution sets $P_l=1-\frac{l}{L+1}$; the \emph{incremental} distribution sets $P_l=\frac{l}{L+1}$; the normal distribution sets $P_l\sim N(0.5, 0.1)$.}\label{fig:}
\end{figure}

Despite the efficiency gains offered by STLD, a unique concern is its sensitivity to the dropout-rate configuration, which involves two key aspects. First, for each device, the average dropout rate across all layers (\ie, $\frac{1}{L}\sum_{l=1}^{L}P_{l}$) significantly impacts both fine-tuning efficiency and model performance. An excessively high dropout rate hinders the LLM’s ability to learn complex patterns, whereas a low dropout rate increases training overhead, thereby slowing the training process (Figure~\ref{fig:impact-average-drop-rate}). Second, within a LLM, the dropout rate for each transformer layer must be carefully set. According to our experiments, different distributions of dropout rates cross layers lead to variations in model accuracy (Figure~\ref{fig:impact-drop-rate-distribution}). Therefore, selecting an "optimal" dropout-rate configuration is critical for achieving both high training efficiency and superior model performance in DropPEFT.
To quantify the utility of a specific configuration, we adopt the widely used time-to-accuracy metric \cite{lai2021oort, li2022pyramidfl, cai2023efficient}, which indicates the wall clock time for training a model to reach a target accuracy. This metric captures the interplay between training efficiency and model performance by revealing how quickly the model accuracy improves over time.

\begin{figure}[t]\centering
    \includegraphics[width=0.67\linewidth]{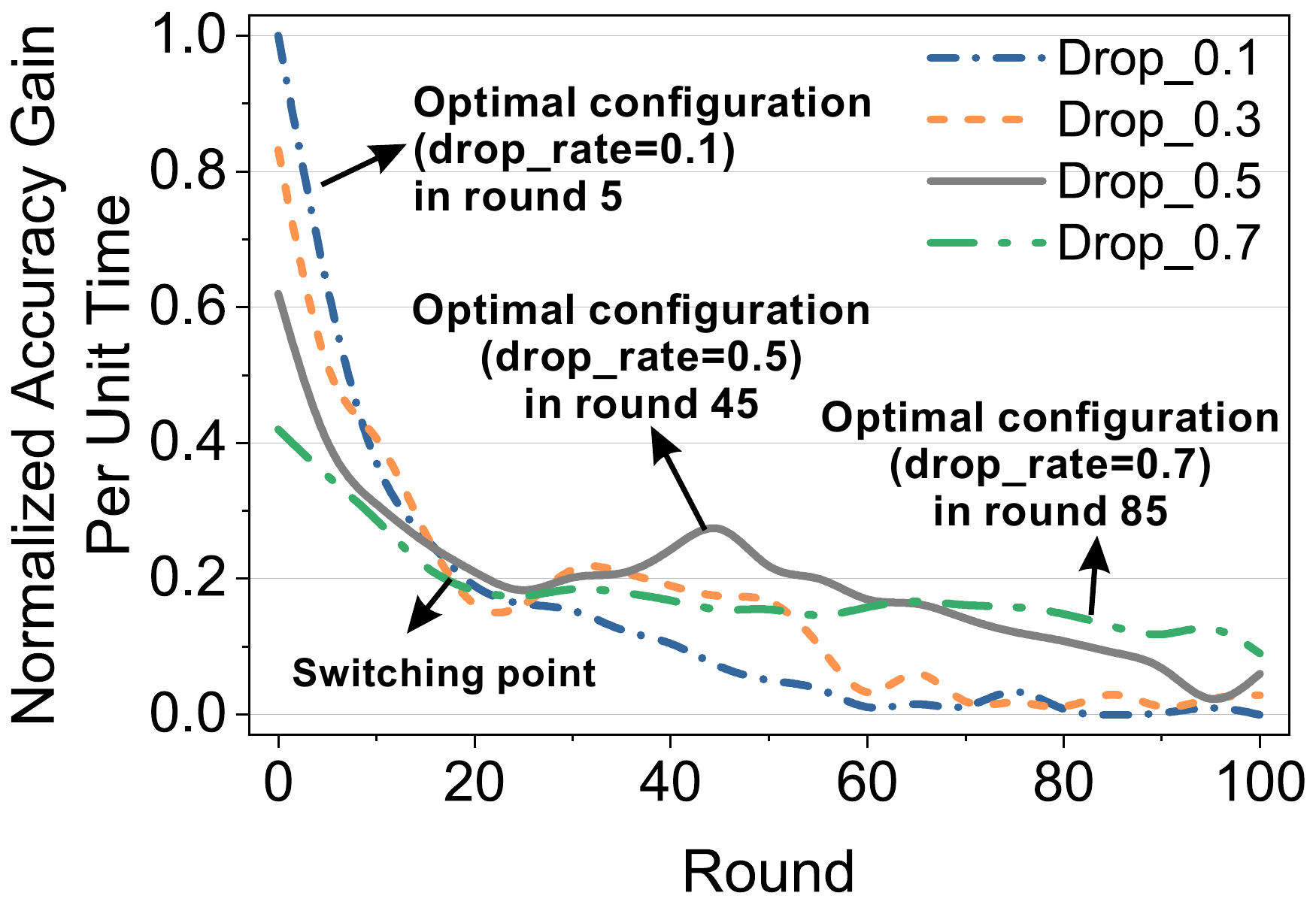}
    \vspace{-2mm}
    \caption{Speeds of the accuracy gains across multiple training rounds under different dropout rate configurations. Model and dataset: RoBERTa-large with MNLI.
    }\label{fig:accuracy_speed}
    \vspace{-5mm}
\end{figure}

\noindent \textbf{Configuration challenges.} Determining an optimal configuration towards high time-to-accuracy performance is challenging. 
First, suitable configurations must adapt to the heterogeneous resources of different devices.
Besides, the desired configuration may change across FL rounds during a fine-tuning session. Figure \ref{fig:accuracy_speed} elaborates how such switching is important for achieving the highest gain in model accuracy per unit time as the training process evolves. Therefore, relying on a fixed configuration determined offline is both difficult and inadequate.


\noindent \textbf{Online exploration-exploitation of configurations.} To tackle these challenges, 
we developed an online configurator that automatically adjusts dropout-rate configurations using an exploration-exploitation strategy (Algorithm \ref{alg:configurator}). This strategy balances exploration (trying new dropout-rate configurations, Line \ref{line:explor_start}-\ref{line:explor_end}) with exploitation (leveraging historically successful configurations, Line \ref{line:exploi_start}-\ref{line:exploi_end}) to converge on near-optimal time-to-accuracy outcomes. 
This process can be modeled as a multi-armed bandit problem \cite{slivkins2019introduction}, where each configuration is an “arm” of the bandit, and the accuracy gains per unit time is the “reward”. Next, we adaptively balance the exploration and exploitation of different arms to maximize long-term rewards, which accumulate to yield high time-to-accuracy performance.


Consider a federated setting with $N$ devices participating in each round, indexed by $i \in \{1, \dots, N\}$. At the beginning of each round, the server selects a dropout-rate vector $\mathbf{P}_i = \{P_{i,1}, \dots, P_{i,L}\}$ for each device $i$ (Line \ref{line:random_start}-\ref{line:random_end} and \ref{line:exploi_start}-\ref{line:narrow}), where $L$ is the number of transformer layers in the LLM. Then, each device performs local fine-tuning by STLD with its assigned dropout rates over multiple batch steps (Line \ref{line:training1}, \ref{line:training2} and \ref{line:training_start}-\ref{line:training_end}).
The fine-tuned LLM's accuracy $A_i$ is evaluated on the local validation set and the wall-clock time $T_i$ for the round is measured as the sum of computation and communication times.
We define the reward $R(\mathbf{P}_i)$ of configuration $\mathbf{P}_i$ on device $i$ as the accuracy gain $\Delta A_i$ per unit time:
\begin{equation}
    R(\mathbf{P}_i) = \frac{\Delta A_i}{ T_i}
\end{equation}
The algorithm aims to identify dropout-rate configurations that maximize the accuracy improvement per time for each device, thereby minimizing the overall time-to-accuracy. For configurations that have not been previously selected, the configurator randomly explores potential options (Line \ref{line:random_start}-\ref{line:random_end}). Meanwhile, the configurator narrows down the decision space by exploiting configurations that have yielded high rewards (Line \ref{line:narrow_start}-\ref{line:narrow_end}, \ref{line:narrow}).
The exploration-exploitation strategy adapts as the training process evolves, letting the algorithm quickly discard underperforming (Line \ref{line:narrow_end}) and overly stale (Line \ref{line:stale}) dropout-rate configurations.


\begin{algorithm}[t]
\footnotesize
\caption{Online Configurator for dropout rates}\label{alg:configurator}
\SetAlgoNoEnd
\KwIn{Target accuracy $Acc_t$; Exploration interval $explor\_r$ (\eg, 5); Exploration rate $\epsilon\in [0, 1]$;
Start-up configuration list $list$; Number of candidate configurations $n$; Configuration window size $size_w$.}

\SetKwFunction{fclient}{ClientTraining}
\SetKwFunction{fserver}{ServerController}
\SetKwProg{Fn}{Function}{:}{}

{\color[RGB]{148,0,211}\scriptsize{\tcc{Initialize configurations.}}}
Candidate configurations $list_{c} \leftarrow list$\; 
Historical configurations $list_h \leftarrow \emptyset$\;

$is\_explore \leftarrow True$\;

\While{$\frac{1}{N} \sum_i^N A_i < Acc_t$}{

    {\color[RGB]{148,0,211}\scriptsize{\tcc{Explorations of configurations.}}}
    \eIf{$is\_explore$ is $True$}{ \label{line:explor_start}
    
        Randomly generate $(n\cdot \epsilon)$ configurations $list_{e}$\;\label{line:random_start}
        
        $list_{c} \leftarrow list_{c} \cup list_{e}$\; \label{line:random_end}
        
        \For{each configuration $\{\mathbf{P}_i\}_{i=i}^{N}$ in $list_{c}$}{
             \fclient{$\{\mathbf{P}_i\}_{i=i}^{N}$}\; \label{line:training1}
             Aggregate received local model updates\;
        }
        $list_{h} \leftarrow list_{h} \cup list_{c}$\;
        $list_{h} \leftarrow$ Latest $size_w$ configurations in $list_{h}$\; \label{line:stale}
        Update rewards for configurations in $list_{h}$\; \label{line:narrow_start}

        $list_{c} \leftarrow$ Configurations in $list_{h}$ with top-$(n\cdot(1-\epsilon))$ rewards\; \label{line:narrow_end}
        
        $is\_explore \leftarrow False$\; \label{line:explor_end}
    }
    {{\color[RGB]{148,0,211}\scriptsize{\tcc{Exploitations of configurations.}}}
        Exploitation round $r \leftarrow 0$\;\label{line:exploi_start}
        $\mathbf{P}_{win} \leftarrow$ Configuration in $list_{h}$ with highest reward\; \label{line:narrow}
        \While{$r < explor\_r$}{
            \fclient{$\mathbf{P}_{win}$}; $r++$\; \label{line:training2}
            Aggregate received local model updates\;
        }
        
        $is\_explore \leftarrow True$\;\label{line:exploi_end}
    }
}
\SetKwProg{Fn}{Function}{:}{}
\Fn{\fclient{$\mathbf{P}$}}{ \label{line:training_start}
    \For{each device $i = 1$ to $N$ \textbf{in parallel}}{
        Send $\mathbf{P}_i \in \mathbf{P}$ to participating device $i$\;
        Train locally by stochastic layer dropout using $\mathbf{P}_i$\;
        Upload local model updates to server\; \label{line:training_end}
    }
}

\end{algorithm}


\noindent \textbf{Further narrowing the decision space.} For practical implementation, we propose two simple yet effective options for reducing the complexity of the decision space. First, developers can discretize continuous dropout rates into a finite set (e.g., $\{0.0, 0.1, 0.2, \dots, 0.9\}$), thereby yielding a finite action space for each configuration.
Additionally, a suitable dropout rate distribution across layers can be preset based on prior experience or pre-experiments. In this way, the online configurator only needs to determine the average dropout rate for all layers on each device rather than specifying the dropout rate for each individual layer. 
In practice, we recommend the incremental distribution, \eg, $P_{l}=\frac{l}{L+1}$,
which has been observed to work particularly well across many models and datasets. This is because early layers extract low-level features
that are subsequently utilized by later layers and therefore should be more reliably preserved.

\section{Adaptation to Heterogeneous Data}\label{sec:personalized}
DropPEFT significantly reduces the overhead of LLM fine-tuning. However, a practical challenge, \ie, statistical heterogeneity, remains in federated fine-tuning tasks, adversely affecting training performance. In real-world scenarios, user data across devices are generated under different contexts, rendering them non-independently and identically distributed (non-IID) \cite{lai2021oort, li2021hermes, zhang2023fedpetuning}.
Conventional federated PEFT frameworks aggregate local model updates from all transformer layers to update a single global LLM. 
While effective in IID settings, this strategy underperforms in non-IID environments.
In this section, we extend DropPEFT to handle non-IID data through a novel approach termed personalized transformer layer sharing (PTLS). This method enables each device to learn customized representations tailored to its local data while concurrently leveraging the shared knowledge derived from the global data of all devices.



\noindent \textbf{Method overview.}
At its core, PTLS enables selective sharing of transformer layers globally while maintaining others as device-specific. The layers in the LLM are divided into \emph{shared} and \emph{personalized} layers. The shared Layers, which remain identical across devices, capture collective language patterns and prevent local LLMs from overfitting to local data. In contrast, the personalized layers are unique to each device, allowing the model to adapt to the specific nuances of local data. 
In communication rounds, each device transmits only the updates from its shared layers to the server for global aggregation.

\noindent \textbf{Selection of shared and personalized layers.} Determining which layers to share versus personalize is crucial for training performance, 
To facilitate adaptive layer personalization across devices, 
We leverage the "gradient criterion" \cite{mortaheb2022fedgradnorm, nguyen2024towards}, which posits that layers exhibiting larger gradients during training are more sensitive and thus more critical for capturing unique data patterns. 
Motivated by this criterion, 
we compute the gradient norm $g_l^{(t)}$ of each layer $l$ for every batch $b$, and then average these values:
\begin{equation}
    I_l = \frac{1}{\sum_{b=1}^{B}(1-d_{l}^{(b)})} \sum_{b=1}^{B} g_l^{(b)}(1-d_{l}^{(b)})
\end{equation}
where $B$ is the total number of batches and $d_{l}^{(b)}$ indicates whether layer $l$ is deactivated or activated by STLD in batch $b$.
A high value of $I_l$ indicates that layer $l$ requires substantial adjustment to adapt to the local data distribution, and should thus be maintained as a personalized layer. Conversely, layers with lower $I_l$ values are more stable and should be incorporated into the global aggregation to share knowledge. In each round, each device uploads updates from $k$ (\eg, $k = L/2$) layers with the lowest $I_l$ values to the server.

\begin{figure}
    \centering
    \includegraphics[width=1.0\linewidth]{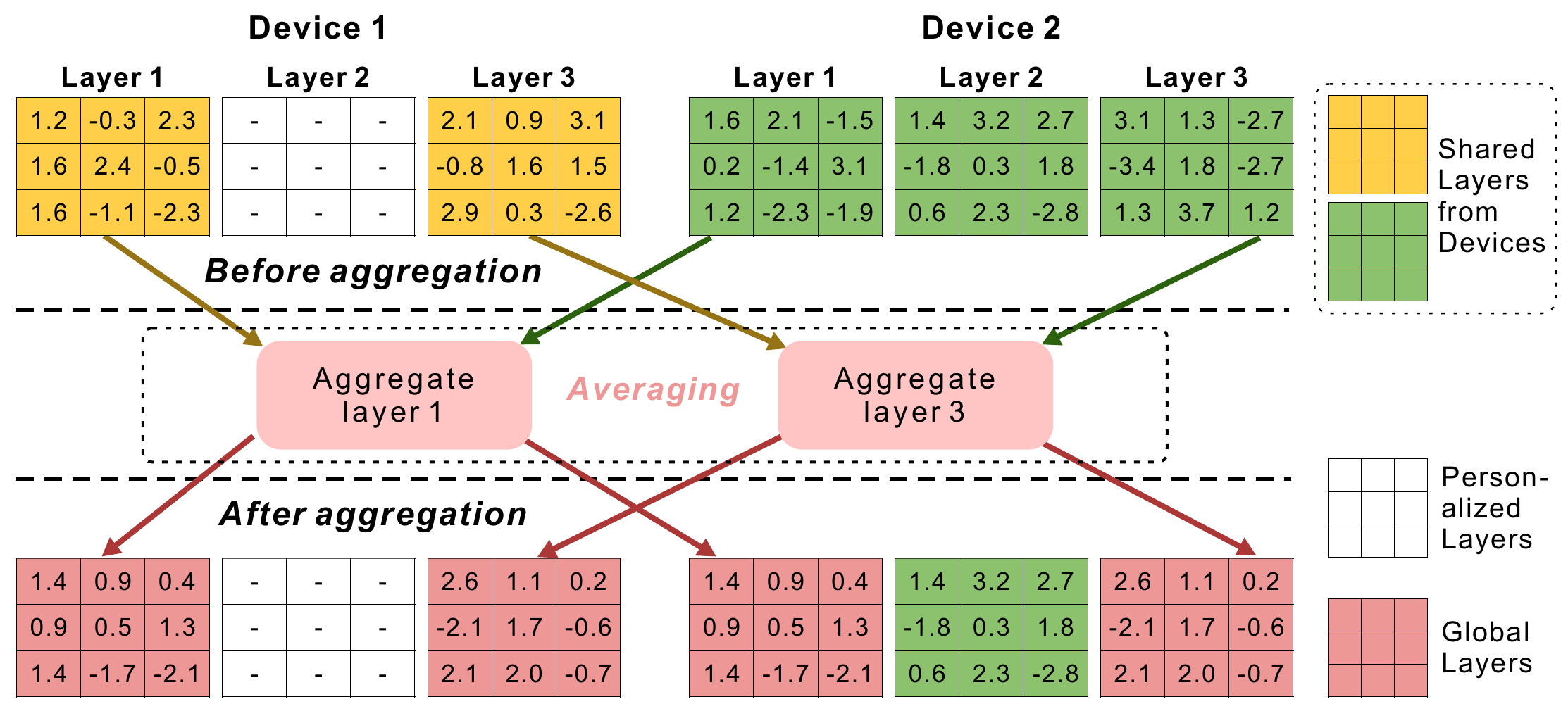}
    \caption{An example of heterogeneous aggregation.}
    \label{fig:hete_agg}
\end{figure}

\noindent \textbf{Heterogeneous layer aggregation.} Considering the non-IID data distribution across devices, their shared layers may be heterogeneous; that is, only parts of these layers overlap among devices.
To handle the heterogeneous layers, we propose an aggregation strategy that averages only the overlapping portions of the layers while keeping the non-overlapping parts unchanged.
As Figure \ref{fig:hete_agg} illustrates, the first and third layers of device $1$ and $2$ overlap, so the parameters of these layers are averaged accordingly. 
Since the second layer of device $1$ is personalized, there is no intersection of this layer between the two devices, and it is therefore excluded from aggregation. Notably, devices holding similar local data will share a greater proportion of overlapping layers. This promotes mutual benefits among devices with similar data distributions while reducing interference among those with different distributions. 

\section{Implementation}
The DropPEFT prototype, which comprises approximately 2,500 lines of Python code, is built atop FedPETuning \cite{zhang2023fedpetuning}, a state-of-the-art benchmark and open-source platform designed for PEFT methodologies. To ensure compatibility with contemporary LLM architectures,
DropPEFT integrates seamlessly with the widely adopted \emph{Transformers} library \cite{Transformers}, leveraging its modular APIs for LLM initialization. To implement the STLD mechanism, we directly modified core LLM modules within the \emph{transformers.models} package \cite{transformers_models}, introducing probabilistic dropout gates that dynamically mask subsets of transformer layers during fine-tuning.
An illustrative implementation example of STLD is provided in Figure \ref{code:stochastic_layer_dropout}. For compatibility with diverse PEFT strategies, \eg, Adapter and LoRA, we integrated the \emph{Opendelta} API \cite{hu2023opendelta}, a plug-and-play library that enables non-invasive injection of trainable PEFT modules (\eg, low-rank matrices and adapters) into frozen pretrained LLMs.
Local fine-tuning workflows are orchestrated using \emph{PyTorch} \cite{paszke2019pytorch}, with CUDA v12.3 and cuDNN v9.1.0 for GPU acceleration.
Distributed communication between devices and the central server is managed through \emph{torch.distributed}, PyTorch's native library for parallel and distributed training, which provides a collection of sending and receiving interfaces for parameter synchronization, \eg, \emph{torch.distributed.send} and \emph{torch.distributed.recv}.

\section{Evaluation}
\subsection{Experimental Setup}

\begin{table}[t]
    \centering
    \caption{Development boards used in experiments.} \label{table:hardware}
    \vspace{-2mm}
    \resizebox{85mm}{!}{
        \centering
        \begin{tabular}{c|c|c|c}
            \hline
            Device & GPU & CPU & Performance\\
            \hline
            TX2 & 256-core Pascal (8GB) & 2-core Denver 2 (64bit) & 2 TFLOPS\\
            NX & 384-core Volta (16GB) & 6-core Carmel (64bit) & 21 TOPS\\
            AGX & 512-core Volta (32GB) & 8-core Carmel (64bit) & 32 TOPS\\
            \hline
        \end{tabular}
    }
    \vspace{-4mm}
\end{table}

\noindent \textbf{Models.} We evaluate DropPEFT mainly on four popular LLMs, \ie, BERT-large \cite{devlin2018bert}, RoBERTa-base \cite{liu2019roberta}, RoBERTa-large \cite{liu2019roberta}, and DeBERTaV3-large \cite{he2021debertav3}. RoBERTa-base contains 12 transformer layers, while BERT-large, RoBERTa-large and DeBERTaV3-large consist of 24 layers. The pre-trained weights for all models are directly downloaded from Hugging Face \cite{huggingface}. These models have been extensively used in prior federated fine-tuning research \cite{cai2023efficient, lin2021fednlp, cai2023federated, zhang2023fedpetuning, xu2024fwdllm}.

\begin{table*}
    \includegraphics[width=1.0\linewidth]{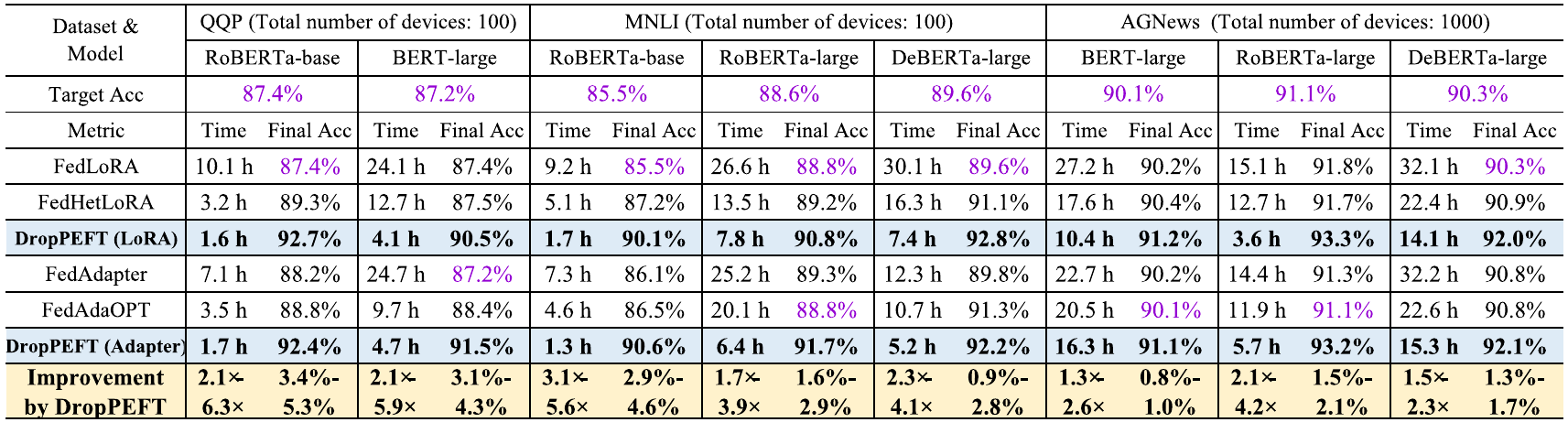}
    \caption{Summary of time-to-accuracy (Time) and final accuracy (Final Acc) of all methods. Time unit: hour (h).}\label{tab:summary_time_acc}
    \vspace{-3mm}
\end{table*}

\noindent \textbf{Datasets.} We conduct experiments on three popular NLP datasets. (1) The Quora Question Pairs (QQP) dataset \cite{wang2018glue} is a collection of over 400K question pairs sourced from the Quora platform. Each pair is labeled to indicate whether the two questions are paraphrases of each other. This dataset is widely used for training and evaluating LLMs on paraphrase identification tasks. (2) The Multi-Genre Natural Language Inference (MNLI) dataset \cite{wang2018glue} is a large-scale collection comprising over 400K sentences used for training and evaluating LLMs on natural language inference (NLI) tasks. In NLI, the goal is to determine whether a premise sentence entails, contradicts, or is neutral with respect to a hypothesis sentence. (3) The AGNews dataset \cite{zhang2015character} is a collection of news articles categorized into four major classes and is commonly used for training and evaluating text classification models. The number of training samples for each class is 30K. These datasets have been extensively leveraged in prior works to validate various PEFT methods \cite{cai2023efficient, cai2023federated, zhang2023fedpetuning, xu2024fwdllm}. Moreover, they are sufficiently large and convenient for federated fine-tuning data partitioning among devices. 

\noindent \textbf{Non-IID data partitioning.} We follow prior literature \cite{zhang2023fedpetuning, cai2023efficient, lin2021fednlp} to divide the datasets across devices using the Dirichlet distribution (100 devices for MNLI and QQP, and 1,000 for AGNews). Specifically, we independently sample training data $\mathcal{D} \sim Dir(\alpha)$ for each device, where $\alpha$ controls the degree of non-IIDness. A lower value of $\alpha$ generates a greater shift in the label distribution. Unless otherwise specified, we use $\alpha = 1.0$ as the default setting throughout our experiments, consistent with FedPETuning. The local test dataset on each device follows a distribution similar to that of the local training dataset.

\noindent \textbf{Hardware.} Consistent with prior FL literature \cite{cai2023efficient, xu2024fwdllm, lai2021oort, li2022pyramidfl, zhang2023fedpetuning}, our experiments are carried out in a semi-emulation manner on an AMAX deep learning workstation with 8 $\times$ NVIDIA A6000 GPUs. On-device training times are measured on three popular end devices (Table \ref{table:hardware}): (1) Jetson TX2 \cite{tx2} is a compact embedded computing platform designed for artificial intelligence applications at the edge;
(2) Jetson NX \cite{nx}, capable of up to 21 TOPS of accelerated computing, delivers the horsepower to run LLMs in parallel. (3) Jetson AGX \cite{agx}, the most powerful of the three, has a computing capability of 32 TOPS. TX2 and NX can work in one of four computational modes, while AGX has eight modes. Devices working in different modes exhibit diverse levels of performance.

\noindent \textbf{Baselines.} We compare DropPEFT with the following baselines: (1) FedAdapter \cite{zhang2023fedpetuning} is a vanilla federated PEFT framework based Adapter. It introduces a small tunable module (\ie, adapter) into each transformer layer while freezing the base LLM.
(2) FedAdaOPT \cite{cai2023efficient}, a state-of-the-art federated Adapter fine-tuning framework, incorporates layer-freezing techniques and a progressive training paradigm to automatically identify the optimal adapter configuration (\ie, adapter width and depth).
(3) FedLoRA, a vanilla federated LoRA fine-tuning framework, interposes low-rank adaptation matrices (\ie, lora modules) into both the multi-head attention module and the feed-forward network in each transformer layer. 
(4) FedHetLoRA \cite{cho2024heterogeneous}, a state-of-the-art federated LoRA fine-tuning framework, allows heterogeneous ranks of lora modules across devices based on their individual system resources. It fine-tunes these modules efficiently through local rank self-pruning and aggregates them by sparsity-weighted aggregation at the server.

For a fair comparison, DropPEFT is implemented on top of LoRA and Adapter, referred to as DropPEFT (LoRA) and DropPEFT (Adapter), respectively.
Notably, DropPEFT adopts the same PEFT configurations (\ie, adapter width/depth and ranks of lora modules) as FedAdapter and FedLoRA. These configurations are based on prior experience and are not optimized by the advanced algorithms in FedAdaOPT and FedHetLoRA.

\noindent \textbf{Metrics.} We primarily report two sets of metrics to evaluate training and runtime performance. (1) Training performance metrics include time-to-accuracy and final accuracy.
We set the target accuracy as the highest \emph{achievable} accuracy by DropPEFT and all baselines \cite{lai2021oort, li2022pyramidfl}. Otherwise, some may never reach that target. Following prior literature \cite{li2021hermes}, we evaluate model accuracy on each device’s test dataset after the federated fine-tuning process is complete, and report the average accuracy across all devices as the final accuracy. (2) Metrics for runtime performance include memory footprint, energy consumption and network traffic on devices.

\noindent \textbf{FL Settings.} Unless stated otherwise, DropPEFT and all baselines use the same set of hyper-parameters as suggested in the FedPETuning benchmark: mini-batch size as 16; local training epoch as 1; learning rate as 5e-4 for RoBERTa-base, 2e-4 for others; max sequence length as 128 for MNLI and QQP, 64 for AGNews \cite{cai2023efficient, xu2024fwdllm}. For MNLI and QQP, we conduct 100 communication rounds and select 10 devices per round by default \cite{zhang2023fedpetuning}. For AGNews, 100 devices are selected per round \cite{xu2024fwdllm}. The network bandwidth for each device fluctuates randomly between 1 Mbps and 100 Mbps, a typical setting for end devices in prior literature \cite{liao2024parallelsfl, liao2024mergesfl}.

\begin{figure*}
    \centering
    \includegraphics[width=1.0\linewidth]{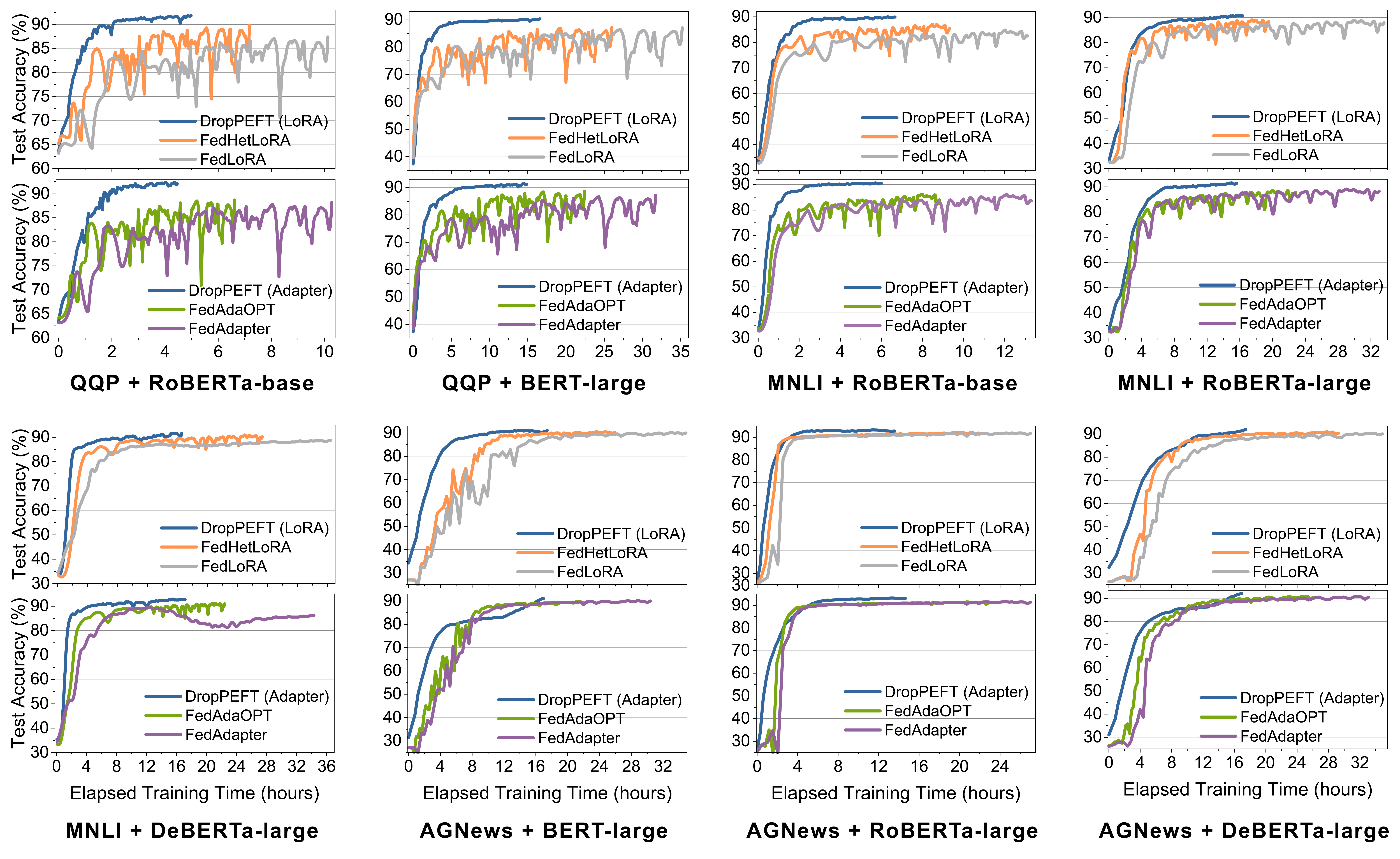}
    \vspace{-5mm}
    \caption{Time-to-accuracy throughout a training session. DropPEFT speeds up model convergence significantly.}
    \label{fig:time_to_acc_e}
\end{figure*}

\subsection{Training Performance}
Table \ref{tab:summary_time_acc} summarizes DropPEFT's improvements on time-to-accuracy and final accuracy over baselines. Figures \ref{fig:time_to_acc_e} reports the timeline of fine-tuning to achieve different accuracy.

\noindent \textbf{DropPEFT improves time-to-accuracy performance.} We notice that DropPEFT speeds up fine-tuning to reach the target accuracy. Specifically, to reach the target on the three datasets, DropPEFT (Adapter) is $1.4$--$5.6\times$ faster than FedAdapter. Besides, the speedup of DropPEFT (LoRA) over FedLoRA is $2.3$--$6.3\times$. The reason is that DropPEFT employs the STLD mechanism, which enables fast fine-tuning by significantly reducing the computational overhead of both the forward and backward passes. Besides, the PTLS method in DropPEFT, which transmits partial layers between devices and the server, notably mitigates the communication time.

More competitive baselines, \ie, FedAdaOPT and FedHetLoRA, only bring limited improvements over FedAdapter and FedLoRA, respectively. FedAdaOPT benefits from an upgrading mechanism on adapter configuration which enables faster boosting of the training accuracy than FedAdapter; FedHetLoRA applies LoRA modules with various ranks to different devices to cater to their heterogeneous system capabilities. However, inherent limitations of FedAdaOPT and FedHetLoRA hinder their improvements: Optimizing configurations of Adapter and LoRA fails to fundamentally solve the problem of high computational overhead in PEFT. In contrast, DropPEFT breaks free from the constraints of existing PEFT methods and optimizes the fine-tuning from a layer dropout perspective. Consequently, DropPEFT (Adapter) takes $1.3$--$3.5\times$ fewer wall clock time on the three datasets to reach the target accuracy than FedAdaOPT. Meanwhile, DropPEFT (LoRA) delivers a speedup of $1.6$--$3.5\times$ over FedHetLoRA.

\noindent \textbf{DropPEFT improves the final accuracy.} 
As represented in Table \ref{tab:summary_time_acc}, DropPEFT consistently outperforms all baselines across various datasets, achieving absolute accuracy gains of $3.1\%$--$5.3\%$, $0.9\%$--$4.6\%$, and $0.8\%$--$2.1\%$ on QQP, MNLI and AGNews, respectively. These improvements are attributed to the PTLS method in DropPEFT. Specifically, in non-IID scenarios, device-level gradients inherently constitute biased estimators of the theoretical global gradient. Baseline methods merge local updates from all transformer layers into a unified global model, thereby amplifying this bias through destructive parameter interference and ultimately driving the global model toward degenerate points where device-specific features are catastrophically forgotten. DropPEFT circumvents this limitation through PTLS, enabling each device to retain personalized layers while collaboratively refining shared parameters. In \S\ref{subsec:component_analysis}, we further quantify the accuracy improvements brought by PTLS under different non-IID settings.


\begin{figure}
    \centering
    \includegraphics[width=1.0\linewidth]{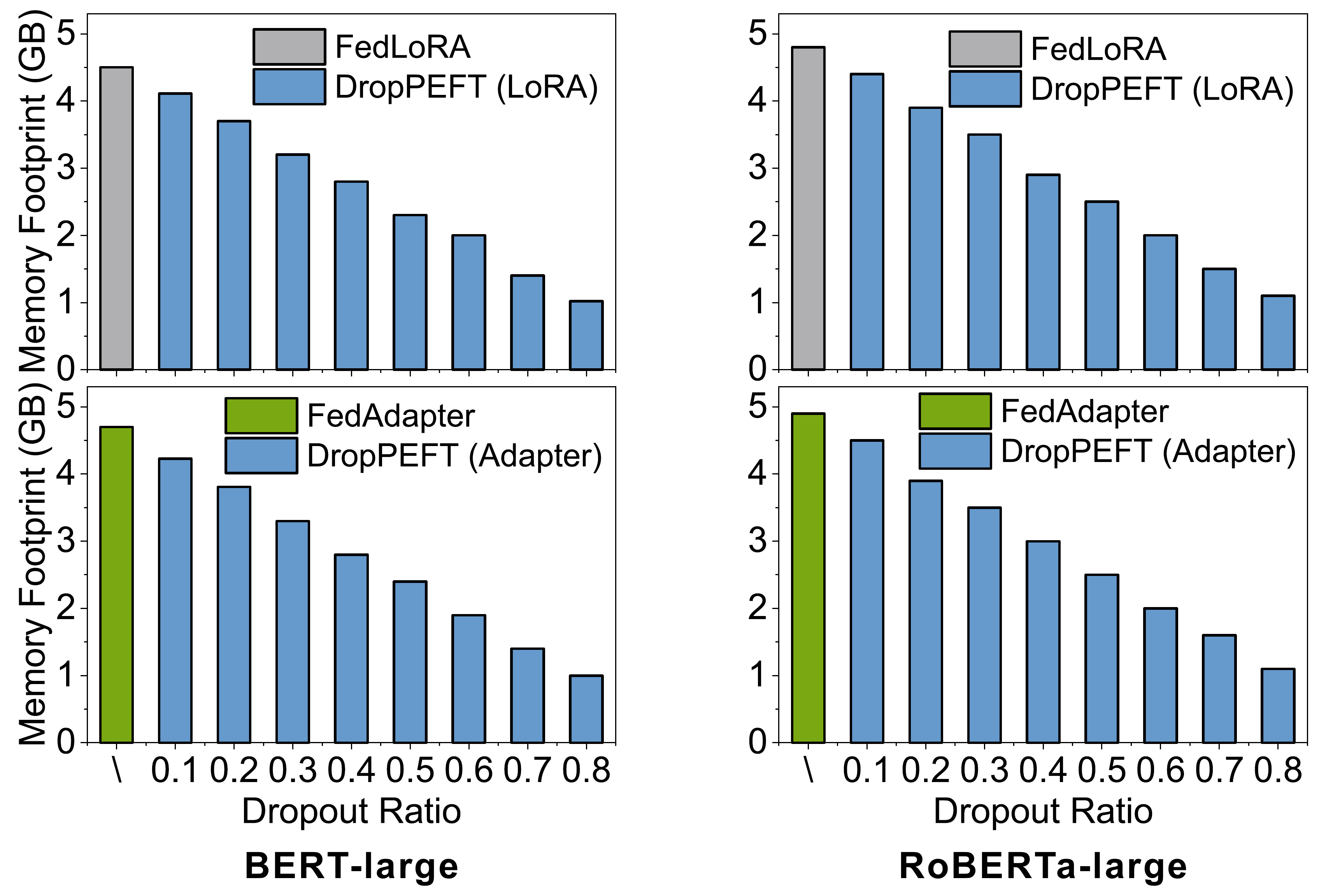}
    \vspace{-5mm}
    \caption{Memory usage of a single device.}
    \label{fig:run_time_memory}
\end{figure}

\subsection{Runtime Performance}\label{subsec:run_time}
We analyze the runtime resource cost during federated fine-tuning, including the peak memory footprint, total energy consumption and network traffic on all devices. The experiments are conducted on the NX device.

\noindent \textbf{DropPEFT reduces the memory usage.} Figure \ref{fig:run_time_memory} reports the peak memory footprint when fine-tuning the BERT-large and RoBERTa-large on AGNews.
DropPEFT nontrivially reduces the memory usage compared to FedAdapter and FedLoRA. For instance, fine-tuning RoBERTa-large with the dropout ratio of $0.6$ reduces the memory footprint over $50\%$ compared to fine-tuning with FedAdapter and FedLoRA. This efficiency stems from the DropPEFT's design, which deactivates a subset of transformer layers during fine-tuning. Therefore, the activations, gradients and optimizer states associated with these layers do not need to be stored. The reduction of the memory usage renders DropPEFT highly suitable for resource-constrained end devices such as TX2 and NX. Moreover, dropout ratios can be dynamically adjusted in each batch of training based on available memory, providing flexibility in adapting to changing device resources.

\begin{figure}
    \centering
    \includegraphics[width=1.0\linewidth]{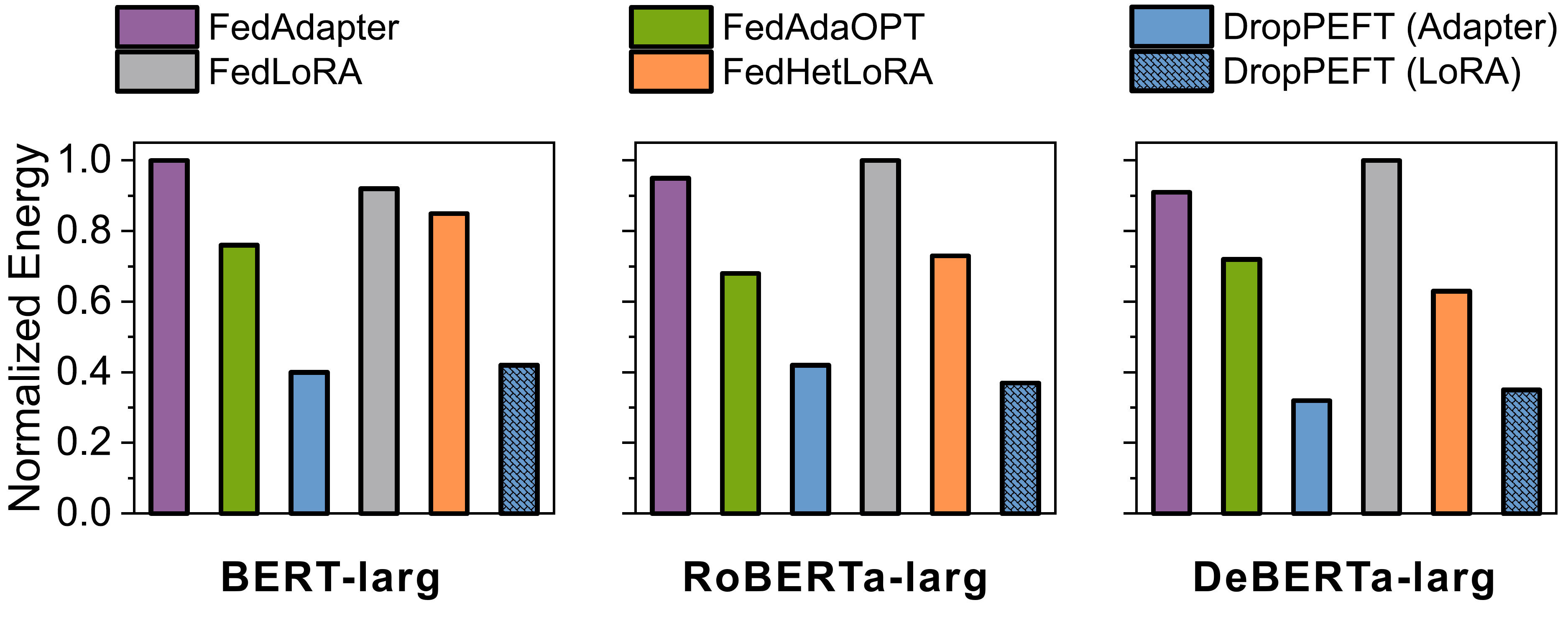}
    \vspace{-0.6cm}
    \caption{Per-device average energy consumption on the MNLI dataset.}
    \label{fig:run_time_energy}
\end{figure}

\begin{figure}
    \centering
    \includegraphics[width=1.0\linewidth]{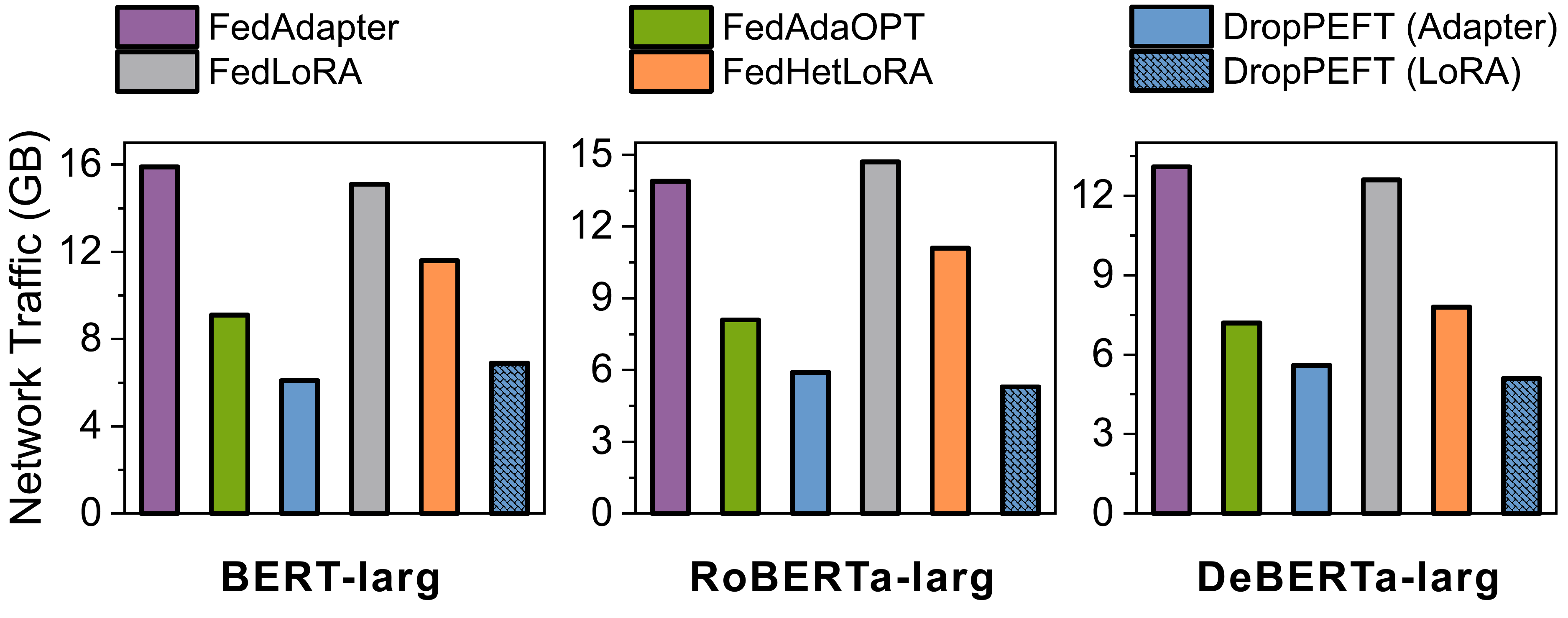}
    \vspace{-0.6cm}
    \caption{Total network traffic of all devices on MNLI.}
    \label{fig:run_time_traffic}
\end{figure}
\noindent \textbf{DropPEFT saves the energy consumption.} Figure \ref{fig:run_time_energy} illustrates the average energy consumed during the federated fine-tuning process on each device. It shows that DropPEFT saves the energy consumption remarkably. Specifically, DropPEFT (adapter) is able to save $55.8\%$--$64.8\%$ and $38.2\%$--$55.6\%$ energy consumption compared to FedAdapter and FedAdaOPT, respectively. Besides, the reduction of energy consumption of DropPEFT (LoRA) over FedLoRA and FedHeLoRA is $56.3$--$60.1\%$ and $44.4\%$--$50.6\%$, respectively. The main reason behind such improvement is that DropPEFT stochastically skips certain layers during fine-tuning, so fewer operations (FLOPs) are performed in each forward and backward pass. Moreover, skipping layers shortens the computation time for each training step. Since energy consumption correlates with runtime, faster iterations reduce total energy use.

\noindent \textbf{DropPEFT mitigates the network traffic.} Figure \ref{fig:run_time_traffic} reports the total network traffic (both uplink and downlink) of all devices incurred during federated fine-tuning to reach the target accuracy. It shows that DropPEFT saves $22.2\%$--$61.6\%$ network traffic compared to the baselines. This is because the devices in DropPEFT only upload and download model updates in a subset of layers to and from the server. The economic implications of these bandwidth reductions prove particularly consequential for commercial federated fine-tuning deployments.


\subsection{Significance of Key Designs}\label{subsec:component_analysis}
The benefits of DropPEFT come from the STLD strategy (\S\ref{sec:layer_dropout}), the automatic configuration for dropout ratio (\S\ref{sec:configurator}), and the PTLS mechanism (\S\ref{sec:personalized}). We now quantify their effectiveness by implementing three breakdown versions.
\begin{itemize}[leftmargin=*]
    \item \textbf{DropPEFT w/o STLD (DropPEFT-b1):} During local fine-tuning, all transformer layers in the LLM are always active.
    \item \textbf{DropPEFT w/o automatic configuration (DropPEFT-b2):} We choose several fixed configurations, \eg, 0.2 and 0.5, through out the federated fine-tuning process.
    \item \textbf{DropPEFT w/o PTLS (DropPEFT-b3):} In each round, all participating devices upload their model updates from all transformer layers to the server for aggregation.
\end{itemize}

\begin{figure}
    \centering
    \includegraphics[width=1.0\linewidth]{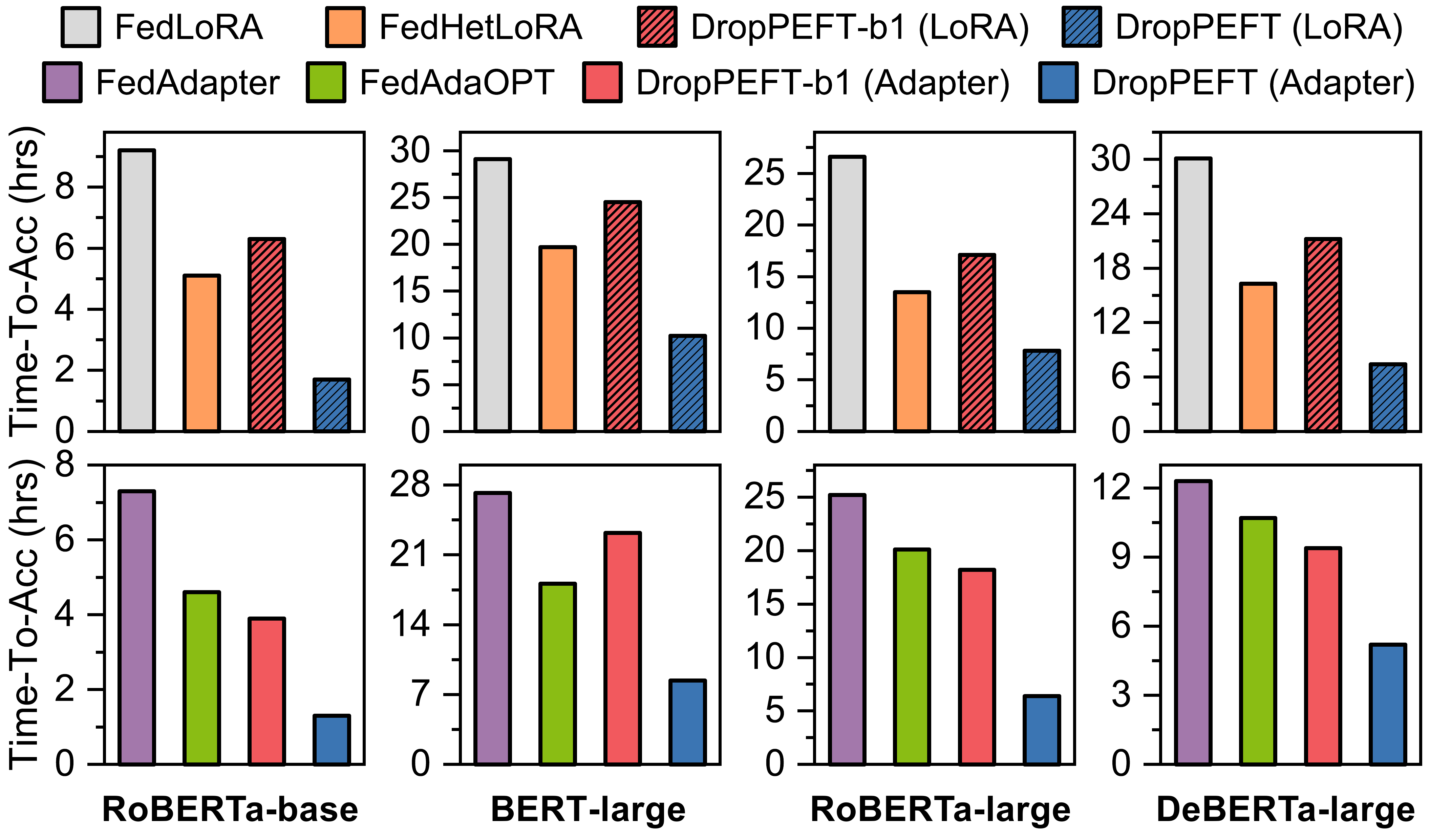}
    \vspace{-0.5cm}
    \caption{Model convergence delays with and without STLD on the MNLI dataset.}
    \label{fig:abl_drop}
\end{figure}

\noindent \textbf{Stochastic transformer layer dropout.} Figure \ref{fig:abl_drop} shows the speedup achieved through STLD. Specifically, removing the layer dropout strategy causes DropPEFT-b1 to revert to the conventional PEFT framework, resulting in convergence delays comparable to FedAdapter and FedLoRA. By employing STLD, DropPEFT brings $1.8\%$--$3.7\%$ speedup compared to DropPEFT-b1. Notably, FedAdaOPT and FedHetLoRA dynamically optimize PEFT configurations (\eg, adapter depth/width or lora rank) during federated fine-tuning to accelerate training. Consequently, DropPEFT-b1, which lacks such configuration optimization, sometimes exhibits slower convergence than FedAdaOPT and FedHetLoRA. Fortunately, STLD is naturally compatible with these optimization methods. This is because layer dropout operates at the granularity of transformer layers, whereas FedAdaOPT and FedHetLoRA modify the PEFT modules \emph{within} each layer without altering the overall model architecture. Therefore, layer dropout can be seamlessly integrated into FedAdaOPT and FedHetLoRA to further improve training efficiency.

\begin{figure}
    \centering
    \includegraphics[width=1.0\linewidth]{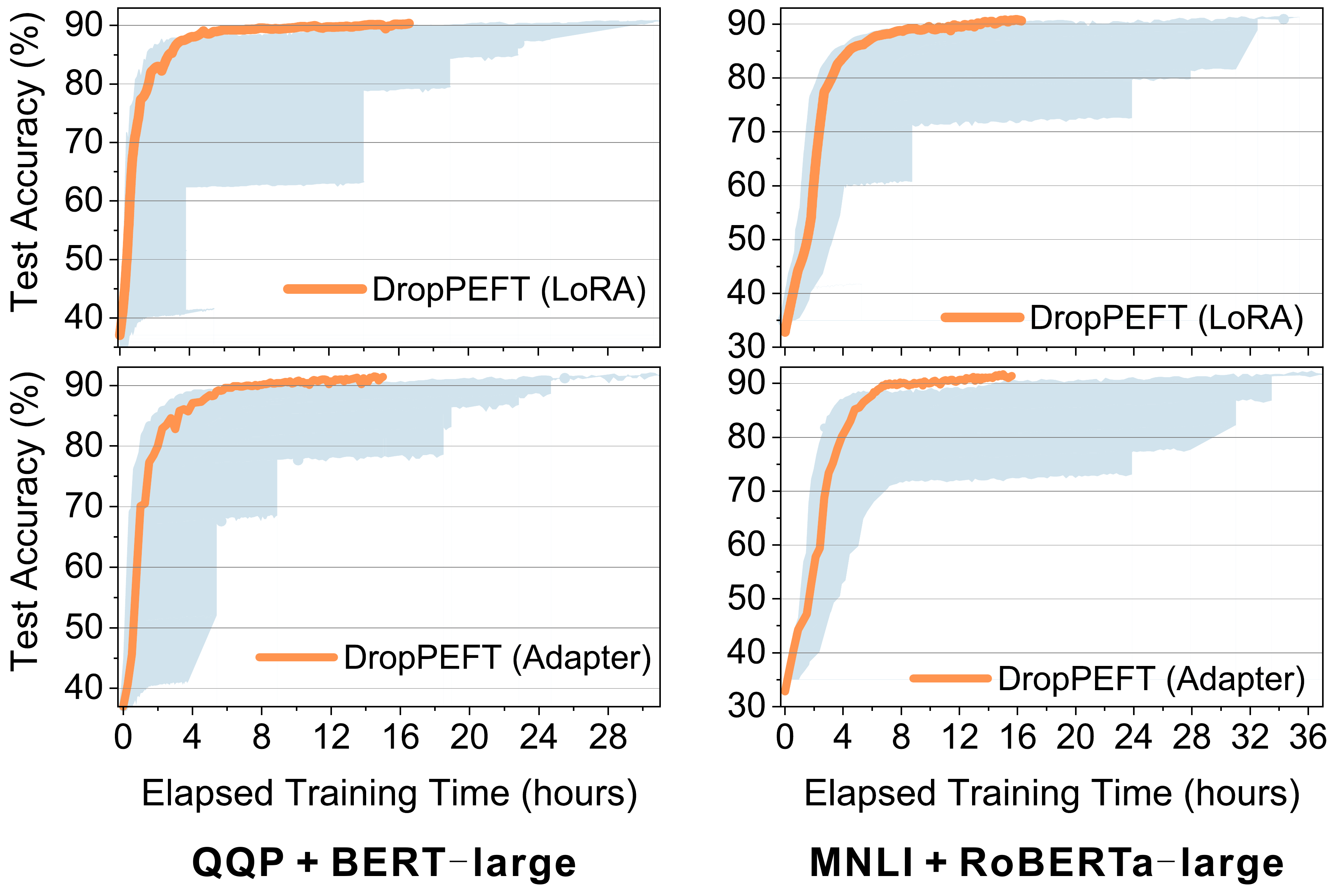}
    \vspace{-0.6cm}
    \caption{Time-to-accuracy throughout a training session.}
    \label{fig:abl_ratio}
\end{figure}

\noindent \textbf{Automatic dropout-ratio configuration.} To demonstrate the importance
of DropPEFT’s adaptively upgrading mechanism on the dropout-ratio configuration, we exhaustively sweep through all fixed dropout-ratio configurations (from 0.1 to 0.9), and aggregate
their convergence curves as blue shaded areas shown in Figure \ref{fig:abl_ratio}. 
The orange line is the curve of DropPEFT.
Note that sweeping all configurations is very expensive, as it takes thousands of GPU hours to run the break versions in a subfigure. The results show that DropPEFT almost outperforms every fixed configuration throughout a training session. This is owing to DropPEFT adaptively selects among different configurations that best suits the current training session.

\begin{figure}
    \centering
    \includegraphics[width=1.0\linewidth]{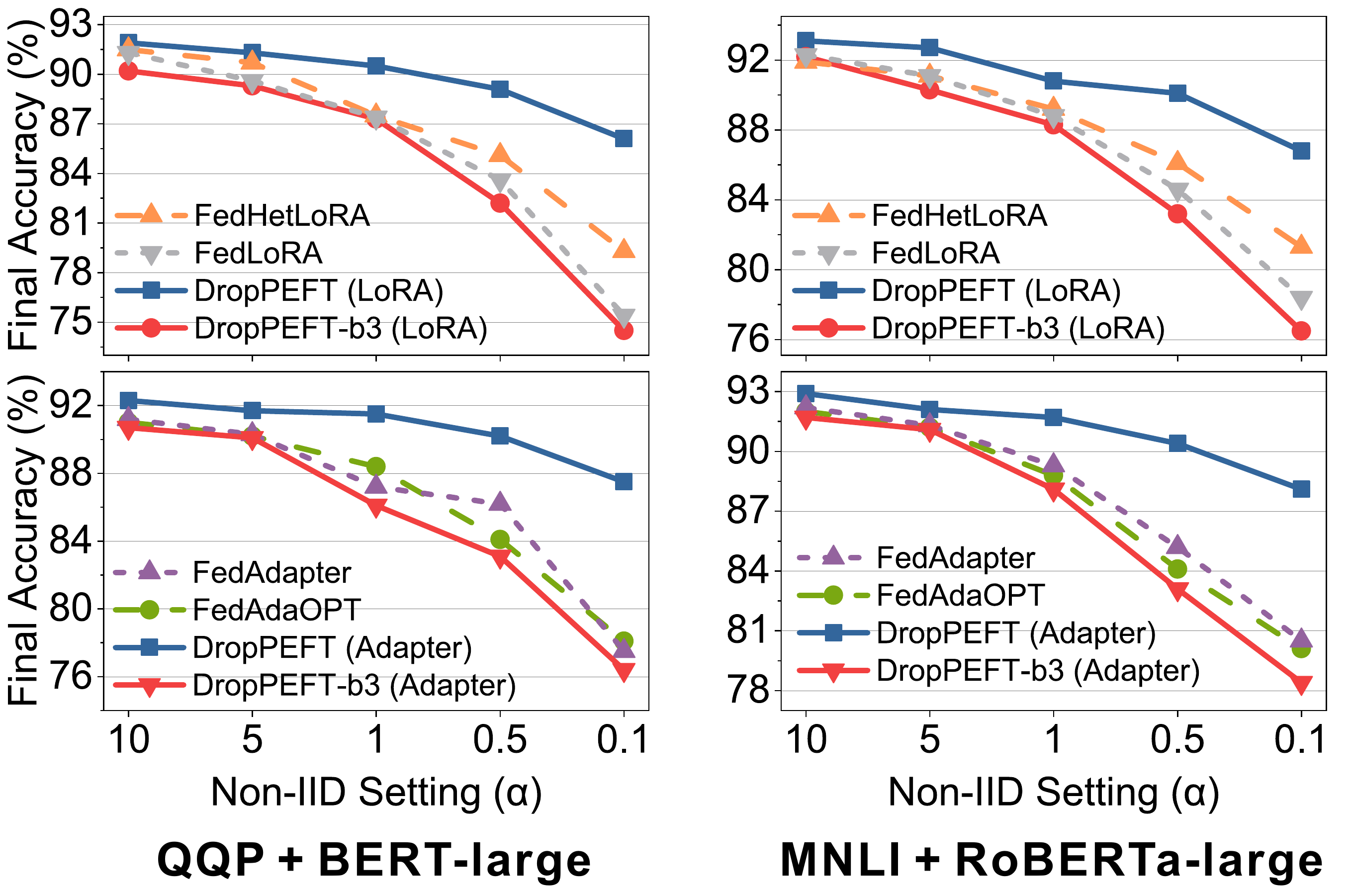}
    \vspace{-0.5cm}
    \caption{Final accuracy under various non-IID settings.}
    \label{fig:abl_pers}
    \vspace{-6mm}
\end{figure}

\noindent \textbf{Personalized transformer layer sharing.} Figure \ref{fig:abl_pers} depicts the final accuracy of different methods under various non-IID settings. Although all methods suffer from model accuracy degradation with a higher degree of non-IIDness, DropPEFT achieves more robust performance compared to other methods. For instance, as the non-IID degree increases (with $\alpha$ decreasing from $10.0$ to $0.1$), final accuracy of DropPEFT-b3 (Adapter), FedAdapter and FedAdaOPT decreases dramatically by $14.3\%$, $13.7\%$, and $12.9\%$ on the QQP dataset, respectively, while DropPEFT has only a $4.8\%$ of accuracy degradation. These results suggest that PTLS enables DropPEFT to effectively improve the final model accuracy by alleviating the negative impact of non-IID data.

\section{Related Work}
\noindent \textbf{Federated fine-tuning of LLMs.} 
While fine-tuning remains the de facto mechanism for adapting LLM to downstream tasks, escalating privacy concerns pose fundamental constraints on centralized data collection from end devices. Federated fine-tuning \cite{wu2024fedbiot, fan2023fate, luo2024fine, wu2024client, hu2024federated} has emerged as a privacy-preserving paradigm for distributed LLM refinement. For instance, the initial exploration, FedNLP \cite{lin2021fednlp}, establishes critical foundations by adapting the classical FedAvg \cite{mcmahan2017communication} framework to construct the first benchmark for federated fine-tuning tasks. Subsequent work, FedPrivate \cite{luo2024fine}, theoretically formalizes the tension between differential privacy guarantees and model utility. Notwithstanding these advancements, the considerable overhead associated with federated fine-tuning remains conspicuously unaddressed.


\noindent \textbf{PEFT methods.} Recent research initiatives increasingly integrate PEFT methodologies \cite{he2021towards, hu2021lora, pfeiffer2020adapterhub, zaken2021bitfit, houlsby2019parameter} into federated fine-tuning frameworks, primarily to address the challenge of communication overhead. The dominant PEFT approaches, \ie, LoRA \cite{hu2021lora} and Adapter \cite{houlsby2019parameter}, achieve communication efficiency by reducing the dimensionality of shared parameters between devices and the server. For instance, FedAdaOPT \cite{cai2023efficient} treats Adapter as a key building block for tackling communication issues in federated fine-tuning, while FeDeRA \cite{yan2024federa} applies truncated singular value decomposition (SVD) to LoRA's update matrices to enhance fine-tuning efficiency. Despite these innovations, our analysis in \S\ref{sec:background} reveals that PEFT-based federated fine-tuning provides negligible relief for on-device training latency and memory pressure. DropPEFT is built atop PEFT but significantly reduces device-side overhead and accelerates convergence by STLD.


\noindent \textbf{FL Optimizations.} 
Substantial research endeavors have sought to optimize cross-device federated learning \cite{mcmahan2017communication, wen2023survey} through various approaches, including communication compression via sparsification or quantization \cite{han2020adaptive, lin2023joint, lang2023joint}, model size reduction by distillation \cite{yang2023fedfed, pang2024federated}, adaptive device sampling strategies \cite{lai2021oort, li2022pyramidfl, mayhoub2024review}, and computational graph optimizations for edge runtime acceleration \cite{zhou2024accelerating, jiang2022model}. While these methods have demonstrated empirical success in classical CNN/RNN architectures, their efficacy diminishes significantly when applied to LLMs. 

\section{Conclusion}
In this work, we propose DropPEFT, an enhanced federated PEFT framework for LLMs that addresses the practical challenges posed by significant overhead in federated fine-tuning. DropPEFT employs a novel STLD method to deactivate a considerable fraction of transformer layers in LLMs during training, thereby eliminating substantial training overhead. To unleash the potential of DropPEFT for high fine-tuning efficiency and training performance, we design an exploration-exploitation strategy that adaptively assigns optimal dropout ratios to devices. Additionally, we extend DropPEFT to handle practical non-IID data by integrating a PTLS approach. Evaluations reveal that DropPEFT outperforms contemporary federated PEFT methods in both fine-tuning efficiency and training performance.

\bibliographystyle{ACM-Reference-Format}
\bibliography{refs}

\end{document}